\documentclass{article}

\PassOptionsToPackage{table}{xcolor}
\usepackage{xcolor}

\usepackage[final]{hallucination_detection}

\usepackage{textcomp}
\usepackage{pifont}
\usepackage{amsmath}
\usepackage[utf8]{inputenc} % allow utf-8 input
\usepackage[T1]{fontenc}    % use 8-bit T1 fonts
\usepackage{hyperref}       % hyperlinks
\usepackage{url}            % simple URL typesetting
\usepackage{booktabs}       % professional-quality tables
\usepackage{amsfonts}       % blackboard math symbols
\usepackage{nicefrac}       % compact symbols for 1/2, etc.
\usepackage{microtype}      % microtypography
\usepackage{xcolor}         % colors
\usepackage{graphicx}
\usepackage{booktabs}  % for better tables
\usepackage{adjustbox} % if you need adjustwidth
\usepackage{tabularx}       % for better table width control
\usepackage[table]{xcolor}  % for row coloring
\usepackage{amssymb}  

\definecolor{lightgray}{RGB}{240,240,240}
\definecolor{cibcred}{RGB}{107,107,107}       % Desaturated: Medium-dark gray (luminance matched)
\definecolor{sectioncolor}{RGB}{107,107,107}  % Same as above
\definecolor{subsectioncolor}{RGB}{58,58,58}  % Slightly adjusted for consistency
\definecolor{subsubsectioncolor}{RGB}{107,107,107} % Same as cibcred
\definecolor{subparagraphcolor}{RGB}{58,58,58} % Slightly adjusted for consistency
\definecolor{lightredforrow}{RGB}{244,244,244} % Very light gray
\definecolor{lighterredforrow}{RGB}{253,253,253} % Nearly white
\definecolor{bordercolor}{RGB}{255,255,255}   % White
\definecolor{tablefontcolor}{RGB}{10,10,10}   % Dark gray
\definecolor{headerred}{RGB}{107,107,107}     % Same as cibcred

\title{Hallucination Detection and Mitigation in Large Language Models}

\author{%
  Ahmad Pesaranghader, Erin Li\\
  CIBC, Toronto\\
  \texttt\{ahmad.pesaranghader, erin.li\}@cibc.com \\
}

\begin{document}

\maketitle

\begin{abstract}
  Large Language Models (LLMs) and Large Reasoning Models (LRMs) offer transformative potential for high-stakes domains like finance and law, but their tendency to hallucinate, generating factually incorrect or unsupported content, poses a critical reliability risk. This paper introduces a comprehensive operational framework for hallucination management, built on a continuous improvement cycle driven by root cause awareness. We categorize hallucination sources into model, data, and context-related factors, allowing targeted interventions over generic fixes. The framework integrates multi-faceted detection methods (e.g., uncertainty estimation, reasoning consistency) with stratified mitigation strategies (e.g., knowledge grounding, confidence calibration). We demonstrate its application through a tiered architecture and a financial data extraction case study, where model, context, and data tiers form a closed feedback loop for progressive reliability enhancement. This approach provides a systematic, scalable methodology for building trustworthy generative AI systems in regulated environments.
\end{abstract}

\section{Introduction}\label{introduction}

Large Language Models (LLMs) and other generative AI systems are revolutionizing industries by automating complex tasks, from summarizing legal documents to generating financial analyses. Their ability to understand and generate human-like text has enabled high-impact applications in finance, healthcare, and legal services, where accuracy and reliability are paramount \cite{zhang2025faith}. However, a critical and persistent limitation of these models is their tendency to \textbf{hallucinate}---that is, to produce outputs that are factually incorrect, semantically inconsistent, or unsupported by the provided context or real-world knowledge \cite{dahl2024legal}.

In high-stakes domains, such as banking systems and legal services, the consequences of hallucination are severe. A financial chatbot misstating insurance coverage limits, or a legal AI misinterpreting a regulatory clause, can lead to significant financial loss, regulatory breaches, and reputational damage \cite{kang2023finance}. The challenge is compounded by the diverse nature of hallucinations stemming from a complex interplay of factors, including model architecture, training data limitations, and inference-time context \cite{magesh2025legalrag}.

\begin{figure}[ht]
    \centering
    \includegraphics[width=0.81\linewidth]{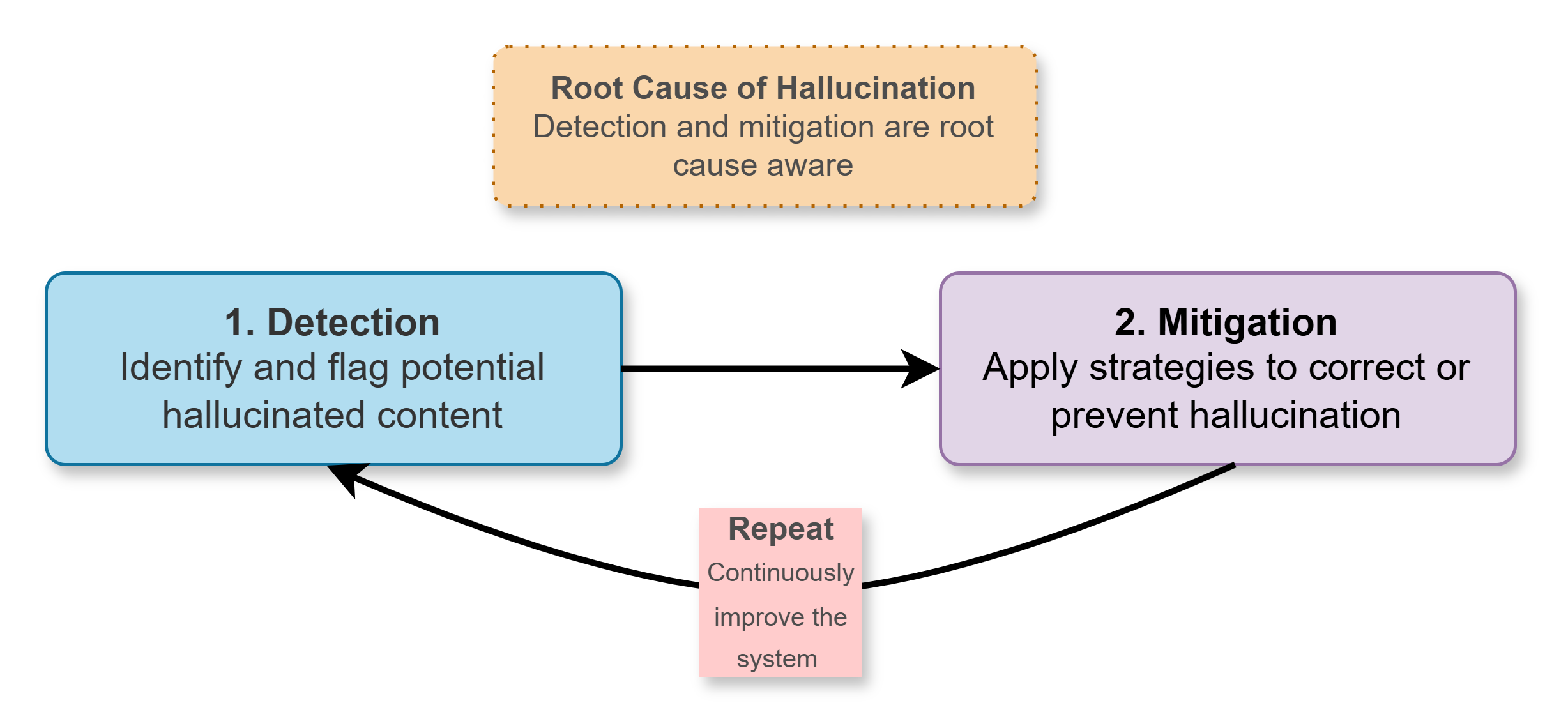}
    \caption{Hallucination Detection \& Mitigation System. This framework implements a continuous improvement cycle where both detection and mitigation strategies are designed to account for potential root causes. The system evolves through iterative testing and refinement.}
    \label{fig1}
\end{figure}

To address this challenge, this paper introduces a structured, operational framework for the systematic management of hallucinations in LLMs and Large Reasoning Models (LRMs). The framework is built on a \textbf{continuous improvement cycle} centered on two core activities: \emph{detection} and \emph{mitigation}. Crucially, the framework is designed to move beyond one-size-fits-all solutions by ensuring both detection and mitigation are \textbf{root cause aware}, as shown in Figure~\ref{fig1}. This diagnostic approach of detection and hallucination is essential because applying a generic fix to a hallucination caused by outdated data (e.g., retraining the model) is inefficient and ineffective compared to a targeted solution (e.g., implementing Retrieval-Augmented Generation).

\begin{figure}[ht]
    \centering
    \includegraphics[width=0.81\linewidth]{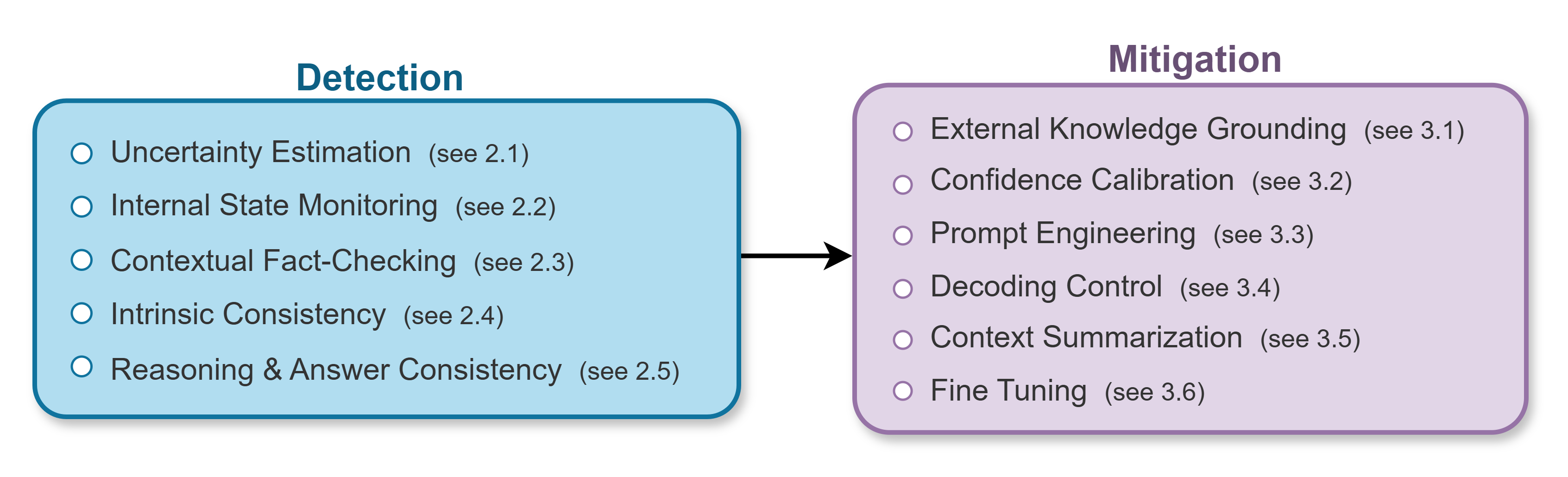}
    \caption{Hallucination Detection Methods \& Mitigation Strategies.}
    \label{fig2}
\end{figure}

This comprehensive framework operates through an integrated, cyclical process where detection and mitigation strategies are systematically aligned based on potential root causes. As visualized in Figure~\ref{fig2}, the system begins with multi-faceted \textbf{Detection}---encompassing uncertainty estimation, factual consistency checks, and reasoning validation---which feeds into targeted \textbf{Mitigation} strategies including knowledge grounding, prompt engineering, and model refinement. This root cause-aware approach ensures diagnostic precision, where specific failure patterns trigger appropriate interventions rather than generic fixes. The continuous cycle evolves through validation and refinement stages, enabling the system to learn from operational evidence and progressively enhance reliability.

This paper details the components of our comprehensive framework. Section~\ref{2-detection-and-diagnosis} covers \textbf{Detection Methods}, where model outputs are evaluated using metrics like uncertainty estimation and factual consistency checks to identify potential hallucinations while maintaining awareness of possible root causes. Section~\ref{3-mitigation-strategies} explores \textbf{Mitigation Strategies}, presenting a toolbox of techniques---from external knowledge grounding to fine-tuning---designed to account for the nature of potential root causes. Section~\ref{4-hallucination-source-hypothesis} delves into the \textbf{Root Cause of Hallucination}, categorizing potential root causes into model, data, and context-related issues to guide system design. Section~\ref{5-operational-framework--continuous-improvement} synthesizes these components into the \textbf{Continuous Improvement Cycle}, explaining how the stages of detection and mitigation form a closed-loop system for enhancing model reliability over time. Section~\ref{6-use-case-application} presents a \textbf{Case Study} applying this framework to document data extraction at scale. The paper concludes with a summary of key findings and future directions in Section~\ref{7-conclusion}.

\section{Detection Methods}\label{2-detection-and-diagnosis}

The \textbf{Detection} stage serves as the operational entry point of the hallucination management framework. Its dual purpose is to (1) confirm whether a hallucination has occurred and (2) maintain awareness of potential root causes. This stage provides \textbf{early warning signals} about reliability risks, and defines the diagnostic pathways that guide precise and cost-effective mitigation. The process begins with detection signals---quantitative or qualitative indicators that an output may be unreliable---and evolves into structured diagnosis, where signals are mapped to potential root causes \cite{dahl2024legal,huang2025survey}.

\subsection{Uncertainty Estimation}\label{uncertainty-estimation}

One of the simplest yet powerful cues in hallucination detection is the model's own internal uncertainty. By analyzing \textbf{token-level probabilities}, \textbf{sequence-level entropy}, \textbf{confidence calibration scores}, or even \textbf{self-declared uncertainty}, systems can identify outputs where the model is unsure.

\subsubsection{Probabilistic vs. Semantic Entropy}\label{probabilistic-vs.-semantic-entropy}

Two complementary entropy modalities are used for hallucination detection:

\paragraph{Log-probability / token-level entropy (probabilistic entropy)}\label{log-probability-token-level-entropy-probabilistic-entropy}

Measures uncertainty in the model's token probability distribution. High token entropy often correlates with weak token-level confidence and can indicate likely hallucination at the surface text level \cite{manakul2023selfcheck}. Token-level entropy is measured by:
\[
H_p = - \sum_{i=1}^{V} p(t_i) \log p(t_i)
\]
where \(p(t_i)\) is the model-assigned probability of token \(t_i\) in a vocabulary of size \(V\).

\textbf{Computational Note:} This calculation requires access to the model's full probability distribution. In \textbf{open-weight} settings, \(H_p\) can be computed directly. In \textbf{closed-weight} or API-based settings where the full distribution is unavailable, \(H_p\) must be \textbf{estimated by sampling multiple generations} and calculating the empirical distribution of the next token.

\textbf{Proxy Measurement:} As a cost-saving alternative for closed-weight models, the entropy of a generated answer can be computed using a separate, open-weight model. However, this measures the \emph{verifier model's} uncertainty about the text, not the original generator's, which can introduce bias if the models differ in knowledge, style, or architecture.

\textbf{Example:} Suppose the model predicts the next token in the sentence \texttt{"The Bank of Canada is expected to \_\_\_ interest rates"} with probabilities: \texttt{"raise"} = 0.6, \texttt{"cut"} = 0.3, \texttt{"hold"} = 0.1. Then, \(H_p = -[0.6\log0.6 + 0.3\log0.3 + 0.1\log0.1] \approx 0.90.\) A lower entropy (e.g., \(0.4\) if one token had \(0.9\) probability) would indicate stronger token-level confidence.

\paragraph{Semantic entropy}\label{semantic-entropy}

Measures uncertainty in \emph{semantic embedding space} rather than token space. This captures uncertainty about meaning and factual grounding beyond surface-token ambiguity, providing better detection of meaning-level inconsistencies \cite{farquhar2024semantic}. Semantic entropy is quantified with the help of:
\[
H_s = - \sum_{k=1}^{K} p(c_k) \log p(c_k)
\]
where \(p(c_k)\) is the normalized probability mass of candidate responses belonging to semantic cluster \(c_k\) among \(K\) clusters in embedding space.

\textbf{Computational Note:} Calculating semantic entropy is inherently more complex than token-level entropy and \textbf{always requires sampling multiple answers} from the model. The standard process is:

\begin{enumerate}
\def\labelenumi{\arabic{enumi})}
\item Sample \(N\) responses to the \textbf{same prompt} using \textbf{stochastic decoding methods} like temperature sampling or nucleus sampling to induce output diversity;
\item Use an embedding model (e.g., a Sentence Transformer) to project responses into a semantic space;
\item Cluster the embeddings (e.g., using agglomerative clustering) to form semantic clusters \(c_k\);
\item Estimate \(p(c_k)\) as the proportion of samples in each cluster.
\end{enumerate}

\textbf{Methodological Considerations:}

\begin{itemize}
\item \textbf{Prompt Design:} For precise fact verification, prompts should target \textbf{single, isolated facts} rather than complex, multi-fact contexts to avoid conflating uncertainties about different pieces of information.
\item \textbf{Sampling Parameters:} The choice of temperature/top-p values requires calibration---excessive diversity can artificially inflate entropy, while insufficient diversity may miss genuine inconsistencies.
\item \textbf{Note on Complex Generations:} For long-form generations containing multiple facts (e.g., summarization), the computed \(H_s\) represents a \emph{global uncertainty} over the entire response. To pinpoint hallucinations to specific claims, the summary must first be decomposed into individual atomic facts, and semantic entropy must be computed for each fact separately (e.g., by generating multiple times for a prompt asking to verify each specific fact).
\end{itemize}

\textbf{Example:} Suppose multiple model responses are generated for the single-fact prompt \texttt{"CIBC's Q2 profit increased compared to Q1."} Four responses confirm the increase (cluster A, \(p(c_A)=0.8\)) and one contradicts it (cluster B, \(p(c_B)=0.2\)). Then, \(H_s = -[0.8 \log 0.8 + 0.2 \log 0.2] \approx 0.50\). A lower entropy (e.g., if all 5 responses agreed, \(H_s=0\)) would indicate high semantic certainty.

\subsubsection{Advanced Uncertainty Estimation Methods}\label{advanced-uncertainty-estimation-methods}

Beyond basic entropy measures, several advanced techniques provide more robust uncertainty quantification:

\paragraph{Monte Carlo Dropout}\label{monte-carlo-dropout}

Monte Carlo (MC) Dropout maintains dropout layers active during inference across multiple stochastic forward passes. Each pass effectively samples from a different subnetwork of the same model, yielding a distribution of predictions. The variance across these predictions estimates epistemic uncertainty, indicating how confident the model is about its internal knowledge \cite{gal2016dropout,kendall2017uncertainties}.

This approach is computationally efficient and straightforward to implement without retraining, making it suitable for high-throughput inference pipelines.

For instance, when extracting financial data from documents (Case Study 1), if the model exhibits high prediction variance for a specific field (e.g., ``interest rate''), it likely hasn't seen similar document patterns during training---flagging it for manual verification.

\paragraph{Ensemble Variance}\label{ensemble-variance}

Ensemble-based methods train or fine-tune multiple models with varying random seeds, architectures, or subsets of training data. The degree of disagreement (variance) among their predictions reflects epistemic uncertainty: if ensemble members disagree widely, the input likely lies outside the familiar data manifold or is inherently ambiguous \cite{lakshminarayanan2017simple,fort2020deep,osband2016deep}.

This method is particularly powerful for open-weight models because they can be replicated, fine-tuned, or perturbed freely---unlike closed-weight APIs where ensemble training is restricted. For example, open-weight models (e.g., LLaMA, Mistral) can easily form ensembles to assess prediction stability or robustness under domain-specific noise.

\paragraph{Bayesian Neural Networks}\label{bayesian-neural-networks}

Bayesian Neural Networks (BNNs) model each parameter (weight and bias) as a probability distribution rather than a fixed value, allowing uncertainty to propagate through the network \cite{blundell2015weight,neal2012bayesian,welling2011bayesian}. This framework captures both:

\begin{itemize}
\item Aleatoric uncertainty (data noise or inherent ambiguity)
\item Epistemic uncertainty (lack of model knowledge)
\end{itemize}

Although computationally intensive, BNNs provide principled uncertainty estimation and can be integrated into hallucination detection pipelines for critical tasks such as structured summarization or financial reporting.

These advanced methods are particularly valuable because they \textbf{do not require calibration with ECE} (Expected Calibration Error)---they provide direct uncertainty estimates that can be used immediately for risk assessment and hallucination flagging.

\subsubsection{Expected Calibration Error (ECE)}\label{expected-calibration-error-ece}

This score quantifies the alignment between a model's predicted confidence scores and its actual empirical accuracy. A perfectly calibrated model would have matching confidence and accuracy---e.g., outputs with 80\% predicted confidence should be correct 80\% of the time.

\textbf{Why ECE matters for hallucination detection}: Traditional uncertainty measures, such as token-level entropy and semantic entropy, are effective at identifying hallucinations where the model exhibits internal uncertainty---for instance, when multiple competing responses have similar probabilities. However, these measures fail completely when LLMs and LRMs produce \emph{high-confidence hallucinations} \cite{huang2025survey,ji2023survey}, where the model consistently generates the same incorrect output with high certainty. In such cases, entropy remains low because the model's probability distribution is sharply peaked around wrong answers. ECE addresses this critical blind spot by quantifying \textbf{miscalibration}---the disconnect between the model's confidence scores and its actual accuracy. By identifying systematic overconfidence, ECE helps flag scenarios where models are \textbf{confidently wrong}, which is essential for risk mitigation in high-stakes domains.

\textbf{Practical implementation:} ECE is computed by binning predictions by confidence, comparing per-bin accuracy with average confidence, and taking the weighted average of these gaps \cite{guo2017calibration}. ECE is measured by:
\[
\text{ECE} = \sum_{m=1}^{M} \frac{|B_m|}{n} \, \big| \text{acc}(B_m) - \text{conf}(B_m) \big|
\]
where \(M\) is the number of confidence bins, \(B_m\) is the set of samples in bin \(m\), \(|B_m|\) its size, \(n\) the total number of samples, \(\text{acc}(B_m)\) the empirical accuracy, and \(\text{conf}(B_m)\) the mean predicted confidence in that bin.

\begin{itemize}
\item \textbf{High ECE} → poor calibration: model's confidence and accuracy diverge (often overconfident).
\item \textbf{Low ECE} → good calibration: model's confidence closely matches actual accuracy.
\end{itemize}

\textbf{Example:} If confidence is divided into 10 bins between 0 and 1.0 (each of width 0.1), a prediction with confidence (0.92) falls into bin 10 \([0.9, 1.0)\). Suppose this prediction is \emph{incorrect} (true label = 0.75). Then, this bin's accuracy may be much lower than its mean confidence (e.g., \(\text{acc}(B_{10})=0.75), (\text{conf}(B_{10})=0.92)\), contributing \(|0.75 - 0.92| = 0.17\) to the overall ECE.

\subsubsection{Self-Declared Uncertainty}\label{self-declared-uncertainty}

Modern LLMs can explicitly \textbf{express confidence} in their own responses through \emph{self-declared uncertainty}---a metacognitive signal where the model communicates how sure it is about an answer. Unlike probabilistic entropy (computed from internal token distributions), self-declared uncertainty is a \textbf{verbalized confidence estimate} derived from the model's own reasoning chain or post-hoc reflection.

\textbf{Why it matters for hallucination detection:} Self-declared uncertainty serves as a complementary signal to probabilistic and semantic entropy. It directly reflects the model's \emph{subjective self-assessment} of reliability, which can be valuable when token-level probabilities are inaccessible or poorly calibrated. Empirical work shows that when prompted to introspect (\texttt{"How confident are you in this answer?"}), models can often predict their correctness above random chance. Integrating this self-assessed confidence helps prioritize or suppress outputs during hallucination screening---particularly in conversational or customer-facing financial applications \cite{kadavath2022language}.

\textbf{Practical implementation:} For each model response, a secondary prompt or reflection step is used to elicit a confidence score, such as:

\begin{quote}
\texttt{"On a scale from 0 to 1, how confident are you that the above statement is factually correct?"}
\end{quote}

\textbf{Example:} When asked \texttt{"Does the Bank of Canada intend to raise interest rates in the 4th quarter of 2025?"}, the model may respond:

\begin{quote}
\texttt{"Yes, it does." (Confidence: 0.65)}
\end{quote}

A relatively low self-declared confidence (\(0.65\)) would flag this output as \emph{uncertain or potentially hallucinated}, guiding retrieval or verification before presentation to users.

\subsection{Internal State Monitoring}\label{internal-state-monitoring}

Beyond self-declared uncertainty, deeper insight is available by probing the \textbf{hidden layers, attention weights, and logit dynamics} of the model. Divergent attention maps or abnormally high logit variance may indicate unstable reasoning trajectories \cite{varshney2024logit}. Monitoring such internal states can reveal hallucinations before they surface in text. For instance, research shows that the attention patterns in specific transformer layers during the ``understanding'' stage of generation are strong indicators of whether the model will subsequently fabricate information \cite{xie2024models}. This method requires tighter integration with the model internals, and thus appeals more to \textbf{in-house AI teams with system access} than external validators.

\subsection{Contextual Fact-Checking}\label{contextual-fact-checking}

To validate outputs against \textbf{external ground truth}, contextual fact-checking compares generations with structured (databases, ontologies) or unstructured sources (retrieved documents). Failures in this comparison usually point to \textbf{outdated training data, knowledge gaps, or missing retrieval alignment} \cite{magesh2025legalrag,lewis2020rag}.

\textbf{Example:} If the model states \emph{``The Bank of Canada's current policy rate is 4.25\%,''} the system retrieves the latest central bank data. If the retrieved rate is \textbf{5.00\%}, the discrepancy triggers a hallucination flag---indicating outdated knowledge or failed retrieval grounding.

\subsection{Intrinsic Consistency (Semantic and Coherence)}\label{intrinsic-consistency-semantic-and-coherence}

Intrinsic evaluation checks the model \textbf{against itself}. This involves testing whether multiple sampled generations are mutually consistent or whether a single long-form answer maintains internal coherence \cite{wang2023selfconsistency}. Contradictions across outputs---e.g., different answers to the same financial query depending on phrasing---signal inference instability or weak reasoning generalization. In operational terms, this is the \textbf{self-consistency test}: outputs that fail it are escalated for stricter validation or mitigation.

\textbf{Example:} When asked,

\begin{enumerate}
\def\labelenumi{\arabic{enumi}.}
\item \texttt{"What was the company's Q2 2024 profit?"} → Model answers \texttt{\$3.7B},
\item \texttt{"How much profit was reported for the second quarter of 2024?"} → Model answers \texttt{\$4.1B},
\end{enumerate}

the contradiction across semantically equivalent queries reveals low intrinsic consistency and a higher risk of hallucination.

\subsection{Reasoning and Answer Consistency Evaluation (RACE)}\label{reasoning-and-answer-consistency-evaluation-race}

The \textbf{RACE framework} extends standard hallucination detection by jointly evaluating ``answer correctness'' and ``reasoning consistency'', penalizing outputs that ``get the right answer for the wrong reasons'' \cite{wang2025race}. This is particularly critical in high-stakes domains where flawed reasoning can undermine trust even if the output appears factually correct. For example, in financial risk assessment, a Large Reasoning Model (LRM) may correctly conclude that a loan application should be rejected or flag a transaction for potential fraud, but cite an incorrect regulatory clause in intermediate reasoning. While the final decision is correct, the flawed reasoning can lead to numerous false-positives and pose verification challenges. Using RACE, such cases can be flagged for review.

\textbf{How it works:} RACE decomposes the joint uncertainty between reasoning \((R)\) and answer \((A)\) given a question \((Q)\) as:
\[
H(R, A \mid Q) = H(R \mid Q) + H(A \mid Q) - I(R, A \mid Q)
\]
where:

\begin{itemize}
\item \(H(R \mid Q)\): uncertainty in reasoning traces,
\item \(H(A \mid Q)\): uncertainty in final predictions,
\item \(I(R, A \mid Q)\): mutual information, reflecting how well reasoning supports the final answer.
\end{itemize}

\textbf{How these are measured:}

\begin{itemize}
\item \textbf{\(H(R \mid Q)\)} is computed as the entropy over reasoning paths generated by the model for the same question---high when reasoning varies significantly across samples.
\item \textbf{\(H(A \mid Q)\)} measures the entropy over final answers across sampled generations---high when the model gives inconsistent outcomes (e.g., ``approve'' vs.~``reject'').
\item \textbf{\(I(R, A \mid Q)\)} is estimated via the correlation or agreement between reasoning and final answers---high when consistent reasoning leads to consistent correct answers, low when reasoning and answers diverge.
\end{itemize}

\textbf{Example:} For the question \texttt{"Should the client's mortgage application be approved under the 2024 affordability guidelines?"}, the model answers \texttt{"No"} (correct), but reasoning cites \texttt{"income-to-loan ratio exceeds 0.7"} instead of the correct \emph{0.6 threshold}. Here, \(A\) is correct but \(R\) is flawed---yielding high \(H(R|Q)\), moderate \(H(A|Q)\), and low \(I(R,A|Q)\). This represents a \textbf{right-answer--wrong-reasoning} case, flagged by RACE for review.

\subsection{Ground Truth Requirements for Detection and Evaluation Metrics}\label{ground-truth-requirements-for-detection-and-evaluation-metrics}

Different hallucination detection metrics have varying dependencies on \textbf{ground truth availability}, which directly impacts their applicability for \textbf{immediate detection} versus \textbf{post-hoc evaluation}. Understanding these dependencies enables the design of a \textbf{multi-layered detection pipeline} that balances \textbf{speed, cost, and accuracy}. Table~\ref{table1} details these requirements.

\begin{table}[ht!]
    \centering
    \small
    \caption{Ground Truth Dependencies Across Hallucination Detection Methods}
    \label{table1}
    \rowcolors{2}{lighterredforrow}{lightredforrow}
    \setlength{\tabcolsep}{10pt}
    \renewcommand{\arraystretch}{1.4}
    \arrayrulecolor{bordercolor}
    \begin{tabularx}{0.95\textwidth}{>{\raggedright\arraybackslash}X>{\centering\arraybackslash}X>{\raggedright\arraybackslash}X}
    \specialrule{1pt}{0pt}{0pt}
    
    \rowcolor{headerred}
    \textbf{\textcolor{white}{Detection Method}} & 
    \textbf{\textcolor{white}{Ground Truth Required?}} & 
    \textbf{\textcolor{white}{Primary Use Case}} \\
    \arrayrulecolor{white}\specialrule{2pt}{0pt}{0pt}\arrayrulecolor{bordercolor}
    
    \textcolor{tablefontcolor}{Uncertainty Estimation} & 
    \textcolor{tablefontcolor}{No (except calibration)} & 
    \textcolor{tablefontcolor}{Real-time screening (entropy/advanced methods); Periodic model calibration (with ground truth)} \\
    \arrayrulecolor{white}\specialrule{1pt}{0pt}{0pt}\arrayrulecolor{bordercolor}
    
    \textcolor{tablefontcolor}{Internal State Monitoring} & 
    \textcolor{tablefontcolor}{No} & 
    \textcolor{tablefontcolor}{Real-time detection with model internals access} \\
    \arrayrulecolor{white}\specialrule{1pt}{0pt}{0pt}\arrayrulecolor{bordercolor}
    
    \textcolor{tablefontcolor}{Contextual Fact-Checking} & 
    \textcolor{tablefontcolor}{Partial (requires reference or retrieved documents)} & 
    \textcolor{tablefontcolor}{Retrieval-augmented generation (RAG) and document-grounded tasks} \\
    \arrayrulecolor{white}\specialrule{1pt}{0pt}{0pt}\arrayrulecolor{bordercolor}
    
    \textcolor{tablefontcolor}{Intrinsic Consistency (Semantic \& Coherence)} & 
    \textcolor{tablefontcolor}{No} & 
    \textcolor{tablefontcolor}{Real-time reasoning stability checks} \\
    \arrayrulecolor{white}\specialrule{1pt}{0pt}{0pt}\arrayrulecolor{bordercolor}
    
    \textcolor{tablefontcolor}{Reasoning \& Answer Consistency (RACE)} & 
    \textcolor{tablefontcolor}{Yes (for training or evaluating CoT alignment)} & 
    \textcolor{tablefontcolor}{High-stakes reasoning and compliance-critical use cases} \\
    
    \specialrule{1pt}{0pt}{0pt}
    \end{tabularx}
\end{table}

\textbf{Notes:}

\begin{itemize}
\item \emph{Uncertainty Estimation metrics} (e.g., token-level entropy, advanced methods) are self-contained and ideal for real-time hallucination monitoring.
\item \emph{Advanced Uncertainty Estimation methods} (Monte Carlo dropout, ensemble variance, Bayesian networks) provide robust uncertainty quantification without requiring ground truth.
\item \emph{Confidence Calibration in Uncertainty Estimation} requires labeled data for measuring deviation between predicted confidence and observed accuracy.
\item \emph{Internal State Monitoring} relies on access to model internals (hidden states, attention maps) and operates without external ground truth.
\item \emph{Contextual Fact-Checking} can often use retrieved context as a weak ground truth.
\item \emph{Intrinsic Consistency} can operate without supervision through internal agreement checking across multiple generations.
\item \emph{Reasoning Consistency (RACE)} depends on ground truth reasoning traces or high-quality reference chains for training or evaluation.
\end{itemize}

\subsection{Open vs. Closed Weight Models for Detection}\label{open-vs.-closed-weight-models-for-detection}

Model transparency fundamentally determines the \textbf{depth and accuracy} of hallucination detection. \textbf{Open-weight models} enable full access to internal activations, facilitating richer uncertainty and reasoning-based detection. \textbf{Closed-weight models} (accessible via APIs) constrain observability, often limiting detection to surface-level outputs. The trade-off lies between \textbf{diagnostic power} and \textbf{operational simplicity}. This is because detection signals by themselves indicate \textbf{the presence} of hallucinations; \textbf{diagnosis} interprets these signals to maintain awareness of \textbf{potential root causes}. Table~\ref{table2} provides detailed pros and cons of a hallucination detection and mitigation system with respect to these models.

\begin{table}[ht!]
    \centering
    \small
    \caption{Detection Capabilities: Open- vs. Closed-Weight Models}
    \label{table2}
    \rowcolors{2}{lighterredforrow}{lightredforrow}
    \setlength{\tabcolsep}{10pt}
    \renewcommand{\arraystretch}{1.4}
    \arrayrulecolor{bordercolor}
    \begin{tabularx}{0.95\textwidth}{>{\raggedright\arraybackslash}X>{\raggedright\arraybackslash}X>{\raggedright\arraybackslash}X}
    \specialrule{1pt}{0pt}{0pt}
    
    \rowcolor{headerred}
    \textbf{\textcolor{white}{Detection Method}} & 
    \textbf{\textcolor{white}{Open-Weight Models}} & 
    \textbf{\textcolor{white}{Closed-Weight Models}} \\
    \arrayrulecolor{white}\specialrule{2pt}{0pt}{0pt}\arrayrulecolor{bordercolor}
    
    \textcolor{tablefontcolor}{Uncertainty Estimation} & 
    \textcolor{tablefontcolor}{Advanced methods (Monte Carlo dropout, ensemble variance, Bayesian heads)} & 
    \textcolor{tablefontcolor}{Restricted to token-level entropy or sampling variance} \\
    \arrayrulecolor{white}\specialrule{1pt}{0pt}{0pt}\arrayrulecolor{bordercolor}
    
    \textcolor{tablefontcolor}{Internal State Monitoring} & 
    \textcolor{tablefontcolor}{Full access to hidden states, attention maps, and logits} & 
    \textcolor{tablefontcolor}{Limited to final text outputs} \\
    \arrayrulecolor{white}\specialrule{1pt}{0pt}{0pt}\arrayrulecolor{bordercolor}
    
    \textcolor{tablefontcolor}{Contextual Fact-Checking} & 
    \textcolor{tablefontcolor}{Can integrate directly with retrieval systems and internal representations} & 
    \textcolor{tablefontcolor}{Limited to output-level comparison with retrieved documents} \\
    \arrayrulecolor{white}\specialrule{1pt}{0pt}{0pt}\arrayrulecolor{bordercolor}
    
    \textcolor{tablefontcolor}{Intrinsic Consistency (Semantic \& Coherence)} & 
    \textcolor{tablefontcolor}{Multiple low-cost inference passes possible for internal consistency} & 
    \textcolor{tablefontcolor}{Feasible but costly and latency-bound via API calls} \\
    \arrayrulecolor{white}\specialrule{1pt}{0pt}{0pt}\arrayrulecolor{bordercolor}
    
    \textcolor{tablefontcolor}{Reasoning \& Answer Consistency (RACE)} & 
    \textcolor{tablefontcolor}{Enables deep inspection of CoT steps and reasoning patterns} & 
    \textcolor{tablefontcolor}{Dependent on whether the model explicitly outputs reasoning traces} \\
    \arrayrulecolor{white}\specialrule{1pt}{0pt}{0pt}\arrayrulecolor{bordercolor}
    
    \textcolor{tablefontcolor}{Custom Detection Models} & 
    \textcolor{tablefontcolor}{Custom classifiers/probes on latent representations can be trained} & 
    \textcolor{tablefontcolor}{No modification; limited to black-box post-hoc analysis} \\
    
    \specialrule{1pt}{0pt}{0pt}
    \end{tabularx}
\end{table}

\textbf{Observation:} Open-weight models provide \textbf{research-grade control and explainability}, while closed-weight models offer \textbf{operational scalability} but limited transparency. Hybrid systems---where closed-weight models are wrapped with open-weight detectors---can sometimes combine both advantages \cite{shorinwa2025survey}. For instance, the generations from a closed-source model (e.g., via API) can be fed as input to a smaller, open-weight model. This detector model can then be used to compute the probabilistic entropy or surprisal of the generated sequence, providing a proxy measure for the closed-weight model's uncertainty without access to its internal states.

\section{Mitigation Strategies}\label{3-mitigation-strategies}

Once hallucinations are \textbf{detected with awareness of potential root causes}, the framework proceeds to mitigation. This stage translates diagnostic insights into \textbf{targeted interventions}, ensuring that corrective measures account for the nature of potential underlying issues rather than applying generic or one-size-fits-all fixes. This stage delivers confidence that hallucination risks are actively managed through systematic controls, and provides a toolbox of \textbf{scalable, source-aligned techniques} that can be tuned and validated over time \cite{zhang2025faith,huang2025survey}.

Mitigation strategies fall into five broad classes: \textbf{external knowledge grounding, confidence calibration, prompt engineering, decoding control, and fine-tuning.} Each strategy is designed to account for specific categories of potential hallucination sources, from knowledge gaps to inference-time instability.

\subsection{External Knowledge Grounding}\label{external-knowledge-grounding}

When potential hallucinations may stem from \textbf{knowledge gaps or data obsolescence}, external knowledge grounding augments the model with \textbf{retrieval-augmented generation (RAG)} or dynamic access to verified databases and APIs \cite{lewis2020rag}. This keeps outputs aligned with up-to-date domain information (e.g., market data, lending policies). It involves building retrieval pipelines (vector databases, hybrid search) and aligning retrieval confidence scores with generation. Advanced methods integrate \textbf{structured sources} (knowledge graphs, ontologies) with unstructured text corpora \cite{magesh2025legalrag}.

This approach reduces hallucinations potentially tied to outdated or incomplete training data while supporting \textbf{verifiable, data-driven reasoning}.

\textbf{Example:} A user asks:

\begin{quote}
\texttt{"What's the current prime lending rate in Canada?"}
\end{quote}

The model retrieves the latest figure from a live financial API or a daily-updated rate database before responding:

\begin{quote}
\texttt{"The prime rate in Canada is 6.95\% as of today."}
\end{quote}

By grounding its output in external data, the model avoids fabricating outdated or incorrect values.

\subsection{Confidence Calibration}\label{confidence-calibration}

A common failure mode, as discussed in the ECE subsection, is \textbf{miscalibrated confidence}---outputs presented with unjustified certainty. Confidence calibration aligns the model's \textbf{predicted confidence with its empirical accuracy}, helping users interpret outputs more reliably.

Calibration techniques adjust predicted probabilities without changing the model's decisions, acting as a \textbf{probabilistic correction layer}. This ensures that, for example, a model prediction with 80\% confidence is actually correct about 80\% of the time.

\subsubsection{Temperature Scaling}\label{temperature-scaling}

Temperature Scaling (TS) is a lightweight \textbf{post-processing} method that calibrates a model by adjusting the sharpness of its output probability distribution. It introduces a single scalar parameter, the temperature \(T\), which is applied to the logits vector \(\mathbf{z}\) before the softmax function \cite{guo2017calibration,pleiss_blog_calibration,adaptive_temp_scaling2024}:

\[
p_i = \frac{\exp(z_i / T)}{\sum_{j=1}^V \exp(z_j / T)}
\]

where \(V\) is the vocabulary size, \(z_i\) is the logit for token \(i\), and \(p_i\) is the resulting calibrated token probability.

\textbf{Core Idea:} The temperature \(T\) acts as a smoothing factor. By scaling the logits, it controls the entropy of the output distribution \textbf{without altering the predicted class ranking}, thus preserving the model's accuracy.

\textbf{How It Works:}

\begin{itemize}
\item The optimal temperature \(T^*\) is \textbf{learned on a validation set} by minimizing the Negative Log-Likelihood (NLL), a proper scoring rule that is minimized when predicted probabilities match the true empirical distribution.
\item \textbf{\(T > 1\) (``Cooling''):} The logits are scaled down (\(z_i/T\) becomes smaller), making the output softmax distribution more uniform (``cooler'' or less confident). This is the common case for reducing overconfidence.
\item \textbf{\(T < 1\) (``Heating''):} The logits are scaled up, making the output distribution more peaked (``sharper'' or more confident).
\end{itemize}

\textbf{Example:} Consider a financial forecasting model that is systematically overconfident:

\begin{itemize}
\item \textbf{Before Calibration:} For inputs where it predicts a stock price increase with \(p = 0.95\), the empirical accuracy is only \(75\%\).
\item \textbf{Calibration:} Using a validation set, we find the optimal temperature \(T^* = 1.5\).
\item \textbf{After Calibration:} The same inputs now yield a calibrated probability: 
\[
p_{\text{calibrated}} \approx \frac{\exp(z_k / 1.5)}{\sum_j \exp(z_j / 1.5)} = 0.78
\] 
This value of \(0.78\) is much better aligned with the observed \(75\%\) accuracy, making the model's confidence more trustworthy.
\end{itemize}

For the uncertainty methods discussed in Sections 2.1.1 and 2.1.2, Temperature Scaling works with token-level probabilities.

\textbf{Temperature Scaling in Probabilistic Entropy}

You apply temperature scaling to the token logits \textbf{before} computing token-level entropy \(H_p\):

\[
H_p^{\text{cal}}(w_t) = -\sum_{v \in V} P_{\text{cal}}(v) \log P_{\text{cal}}(v)
\]

where

\[
P_{\text{cal}}(v) = \frac{\exp(z_t^v / T)}{\sum_{v' \in V} \exp(z_t^{v'} / T)}.
\]

\textbf{Effect:} Scaling by \(T>1\) smooths token probabilities, increasing entropy for overconfident predictions and improving correlation between entropy and true uncertainty.

\textbf{Temperature Scaling in Semantic Entropy}

Temperature scaling affects the probability of entire \textbf{generated sequences} \(x = (w_1, w_2, \dots, w_L)\), used to compute semantic cluster probabilities \(P(S_k)\):

\[
P_{\text{cal}}(x) = \prod_{t=1}^{L} \frac{\exp(z_t^{w_t} / T)}{\sum_{v \in V} \exp(z_t^v / T)}.
\]

Then, as usual:

\[
P_{\text{cal}}(S_k) = \sum_{x \in S_k} P_{\text{cal}}(x),
\quad
H_s^{\text{cal}} = -\sum_{k=1}^{K} P_{\text{cal}}(S_k) \log P_{\text{cal}}(S_k).
\]

\textbf{Effect:} The calibrated probabilities prevent a single cluster (e.g., one semantic interpretation) from dominating due to overconfident logits, producing a more balanced entropy signal that reflects semantic disagreement among sampled outputs.

\textbf{Temperature Scaling in Monte Carlo (MC) Dropout}

Temperature scaling is applied \textbf{independently} to each stochastic forward pass:

\[
p_i^{\text{cal}} = \frac{1}{M} \sum_{m=1}^{M} \text{softmax}\left(\frac{\mathbf{z}_m}{T}\right)_i
\]

where \(M\) is the number of passes and \(\mathbf{z}_m\) the logits from pass \(m\).

\textbf{Effect:} Scaling the logits in every pass adjusts within-pass confidence, often increasing inter-pass variance for overconfident models --- yielding a more realistic measure of epistemic uncertainty.

Temperature Scaling is particularly popular because it adds minimal complexity (one parameter) while often significantly improving calibration, especially for overconfident modern neural networks.

\subsubsection{Isotonic Regression}\label{isotonic-regression}

Isotonic Regression is a powerful \textbf{non-parametric} calibration method that learns a flexible mapping from an uncalibrated classifier's confidence scores to well-calibrated probabilities. Unlike Temperature Scaling, which assumes a fixed parametric form (a simple scaling), Isotonic Regression makes no such assumption---it learns a \textbf{monotonic, piecewise-constant function} directly from the validation data \cite{guo2017calibration,zadrozny2002transforming,niculescu2005predicting}.

\textbf{Core Idea:} Find a monotonic (non-decreasing) function \(f\) that transforms the model's raw scores such that the \textbf{transformed scores match the empirical accuracy} of the corresponding bins as closely as possible.

\textbf{How It Works:}

The algorithm works in two main steps:

\begin{enumerate}
\def\labelenumi{\arabic{enumi}.}
\item \textbf{Sorting and Binning:} Given a validation set of model confidence scores \(s_i\) and their true binary outcomes \(y_i \in \{0,1\}\), the pairs \((s_i, y_i)\) are sorted by the confidence score \(s_i\).
\item \textbf{Pooling Adjacent Violators (PAV):} The Isotonic Regression algorithm (implemented via PAV) then groups these sorted scores into contiguous bins. It finds the function \(f\) that minimizes the sum of squared errors:
\[
\min_{f} \sum_{i=1}^{n} (y_i - f(s_i))^2
\]
subject to the constraint that \(f(s_i) \leq f(s_j)\) whenever \(s_i \leq s_j\) (the monotonicity constraint).
\end{enumerate}

The result is a \textbf{calibration map} where each input confidence score \(s\) is replaced by the average empirical accuracy of its bin, \(f(s)\).

\textbf{Example:}

Here is the underlying process:

\begin{itemize}
\item \textbf{Validation Data:} The model predicts on a validation set. We observe that all instances where it predicted \textasciitilde0.6 confidence have an actual accuracy of 50\%, and all instances with \textasciitilde0.9 confidence have an actual accuracy of 85\%. This reveals the miscalibration.
\item \textbf{Fitting the Function:} Isotonic Regression fits a step function \(f\) to this data. For all raw scores in the range that includes 0.6, it assigns \(f(0.6) = 0.5\). For all raw scores in the range that includes 0.9, it assigns \(f(0.9) = 0.85\).
\item \textbf{Application:} At inference time, when the model outputs a raw confidence score of \texttt{0.6} for a new input, the Isotonic Regression calibrator \textbf{remaps it to \texttt{0.5}}. A score of \texttt{0.9} is remapped to \texttt{0.85}.
\end{itemize}

This method is particularly effective when the miscalibration is severe and non-linear, as it can learn complex correction patterns without being constrained by a simple shape like the sigmoid in Platt Scaling.

\subsubsection{Bayesian Post-hoc Calibration}\label{bayesian-post-hoc-calibration}

Standard neural networks output a single, fixed probability for a given input, which often fails to capture the model's own \textbf{epistemic uncertainty}---its uncertainty due to a lack of knowledge, especially for inputs different from its training data. \textbf{Bayesian Post-hoc Calibration} addresses this by treating the model's parameters as a distribution rather than fixed values. This approach produces a \textbf{posterior predictive distribution}, which inherently represents uncertainty in a more principled way \cite{blundell2015weight,neal2012bayesian,vadera2021posthoc}.

\textbf{Core Idea:} Instead of a single prediction from one set of weights, we make predictions using \emph{many different sets of weights} sampled from the posterior distribution over parameters. The variation in these predictions quantifies the model's uncertainty.

\textbf{How It Works:}

A standard model gives you one answer: \(p(\text{class} \mid \text{input}, \mathbf{w})\), where \(\mathbf{w}\) is the single, best set of trained weights.

A Bayesian neural network gives you a distribution of answers by marginalizing over the possible weights:
\[
p(\text{class} \mid \text{input}, \mathcal{D}) = \int p(\text{class} \mid \text{input}, \mathbf{w}) \, p(\mathbf{w} \mid \mathcal{D}) \, d\mathbf{w}
\]

Since this integral is intractable, we approximate it using techniques like \textbf{Monte Carlo Dropout} or \textbf{Ensembling}:

\begin{enumerate}
\def\labelenumi{\arabic{enumi}.}
\item \textbf{Monte Carlo Dropout:} At inference time, dropout is kept active. The model is run multiple times (\(T\) passes) on the \emph{same input}, with different neurons randomly dropped each time. This produces \(T\) slightly different predictions.
\item \textbf{Deep Ensembles:} Multiple models are trained independently from different random initializations. Predictions from all models are aggregated.
\end{enumerate}

In both cases, the result is not a single probability, but a \textbf{distribution of probabilities} for the given input.

\textbf{Example:}

The example \texttt{"Loan approval probability = 80\%"} is the output of a standard model. A Bayesian post-hoc approach would work as follows:

\begin{itemize}
\item \textbf{Query:} A single loan application.
\item \textbf{Process:} The model (with dropout enabled) is run 100 times on the same application.
\item \textbf{Raw Outputs:} Each run outputs a slightly different probability, e.g., \texttt{[0.82, 0.79, 0.65, 0.88, 0.74, ...]}.
\item \textbf{Calibrated Output:} Instead of a single number, we compute statistics over this distribution of probabilities:
\begin{itemize}
\item \textbf{Mean:} \texttt{80\%} (the average probability, which might be the same as the standard model's point estimate).
\item \textbf{95\% Credible Interval:} \texttt{[65\%, 90\%]} (this is the crucial addition, showing the range of likely probabilities).
\end{itemize}
\end{itemize}

This final output is far more informative. The wide interval from 65\% to 90\% indicates high \textbf{model uncertainty}---the model is ``unsure of its own answer'' because the input is ambiguous or out-of-distribution. A narrow interval (e.g., \texttt{[78\%, 82\%]}) would indicate high \textbf{model confidence} in its 80\% estimate. This nuanced view of uncertainty is what makes the output ``calibrated'' and more trustworthy for high-stakes decisions like loan approval.

\subsubsection{Multi-pass Self-evaluation}\label{multi-pass-self-evaluation}

Traditional calibration methods like Temperature Scaling require a separate, labeled validation set. For LLMs, especially in open-ended tasks, such data is not always available. \textbf{Multi-pass Self-evaluation} addresses this by having the model critique its own generations, using the consistency across multiple critiques as a proxy for confidence \cite{manakul2023selfcheck,kadavath2022language,welleck2024revisiting}.

\textbf{Core Idea:} Instead of a single, deterministic answer, the model generates multiple stochastic reasoning paths and/or self-evaluations. The agreement (or disagreement) across these passes is used to quantify the final confidence score, directly tying the model's uncertainty to its internal consistency.

\textbf{Process Breakdown:} The ``multi-pass'' procedure typically involves two distinct phases:

\begin{enumerate}
\def\labelenumi{\arabic{enumi}.}
\item \textbf{Initial Answer Generation (The ``Regular'' Pass):} The model is given a query and produces a primary answer. This can be a zero-shot response or, more commonly, a Chain-of-Thought (CoT) response that includes its reasoning.
\item \textbf{Self-Evaluation Passes (The ``Critique'' Passes):} The model is then prompted, multiple times with added randomness (e.g., high temperature), to act as a ``critic.'' In each pass, it is asked to evaluate its \emph{initial answer}. The prompt might ask: ``Is the provided answer correct? Justify your reasoning.'' or ``What is the probability that this answer is true?''
\end{enumerate}

The final confidence is not the probability from the first pass, but an aggregate of the self-evaluation outcomes (e.g., the fraction of passes that deemed the answer correct).

\textbf{Example:} The original example, \texttt{"Will interest rates rise next quarter?"}, can be broken down as follows:

\begin{itemize}
\item \textbf{Query:} ``Will interest rates rise next quarter?''
\item \textbf{Initial Answer (Regular Pass):} The model generates: ``Yes, based on current inflation trends and central bank guidance.''
\item \textbf{Self-Evaluation Passes (Three Critique Passes):}
\begin{itemize}
\item The model is prompted three separate times with: \texttt{"Review the following answer and determine if it is correct: '[Initial Answer]'"}.
\item \textbf{Pass 1 Output:} \texttt{"The reasoning is sound. **Yes.**"}
\item \textbf{Pass 2 Output:} \texttt{"The guidance is clear, but external shocks could change this. **Yes.**"}
\item \textbf{Pass 3 Output:} \texttt{"This is highly uncertain; the economic data is mixed. **Unsure.**"}
\end{itemize}
\item \textbf{Aggregation:} The results are aggregated. Two ``Yes'' votes and one ``Unsure'' yield a calibrated confidence of \textbf{\textasciitilde66\%}, accurately reflecting the model's internal disagreement, unlike the potentially overconfident 95\%+ that might have been output by the initial pass alone.
\end{itemize}

This method effectively separates the ``generator'' from the ``critic,'' allowing the model to surface its own uncertainty, which is then used for calibration.

Confidence calibration serves both as an \textbf{early-warning mechanism} and as a \textbf{probabilistic trust layer}. By aligning model confidence with reality, calibrated models reduce hallucinations and support reliable decision-making in financial and analytical systems---where unjustified certainty can lead to costly errors.

\subsection{Prompt Engineering}\label{prompt-engineering}

Prompt engineering mitigates hallucinations stemming from \textbf{vague or ambiguous inputs} by strategically structuring instructions to guide models toward factual, verifiable responses.

\subsubsection{Chain-of-Thought Prompting}\label{chain-of-thought-prompting}

Forces the model to articulate its reasoning process before answering, exposing flawed logic and reducing unsupported conclusions \cite{kojima2022cot}.

\textbf{Core Idea:} Require the model to \texttt{"show its work."}

\textbf{Example:}

\begin{itemize}
\item \textbf{Prompt:} \texttt{"Explain step by step: If Bank A's rate is 3\% and Bank B's is 4\%, which is higher?"}
\item \textbf{Output:} \texttt{"Bank A = 3\%, Bank B = 4\%. 4\% > 3\%. Therefore, Bank B's rate is higher."} → The explicit steps prevent calculation errors.
\end{itemize}

\subsubsection{Few-Shot Exemplars}\label{few-shot-exemplars}

Provides concrete \textbf{example question-answer pairs} to demonstrate the desired format, style, and factuality.

\textbf{Core Idea:} Demonstrate the expected behavior through examples.

\textbf{Example:}

\begin{itemize}
\item \textbf{Prompt:} \texttt{"Q: What is an overdraft? A: A negative balance from exceeding deposits. Q: What is a mortgage? A: A long-term loan secured by property. Q: What is a credit line? A:"} → The model adheres to the established factual, concise pattern.
\end{itemize}

\subsubsection{Instruction Layering}\label{instruction-layering}

Combines multiple explicit constraints (on content, format, and style) into a single, comprehensive prompt.

\textbf{Core Idea:} Leave no room for improvisation with layered commands.

\textbf{Example:}

\begin{itemize}
\item \textbf{Prompt:} \texttt{"Summarize the report below in 3 bullet points. Use only verifiable data from the source. Do not speculate."} → The layered commands (\texttt{"only verifiable data,"} \texttt{"do not speculate"}) enforce grounded outputs.
\end{itemize}

\subsubsection{Self-Consistency Prompting}\label{self-consistency-prompting}

Generates multiple answers to the \textbf{identical prompt} using stochastic sampling (e.g., with a moderate temperature \textgreater{} 0 like 0.5) and selects the most frequent result. This filters out inconsistent outliers that arise from flawed reasoning paths \cite{wang2023selfconsistency}.

\textbf{Core Idea:} A fact is more trustworthy if the model arrives at it consistently through \emph{different stochastic reasoning processes}, not just by repeating the same deterministic path.

\textbf{Why Temperature \textgreater{} 0?} Using a non-zero temperature forces the model to \textbf{explore multiple high-likelihood reasoning sequences}. A correct, well-grounded answer will emerge as a robust \textbf{consensus} across these variations, while hallucinations will appear as inconsistent outliers.

\textbf{Example:}

\begin{itemize}
\item \textbf{Identical Prompt (5 runs, T=0.5):} \texttt{"What was the Q3 profit margin in the report?"}
\item \textbf{Outputs:}
\begin{itemize}
\item Run 1: \texttt{"The profit margin was 18.5\%."}
\item Run 2: \texttt{"It was 18.5\% according to the document."}
\item Run 3: \texttt{"18.5\%."}
\item Run 4: \texttt{"The report indicates 22\%."} ← \textbf{Outlier (low-probability path)}
\item Run 5: \texttt{"18.5\%."}
\end{itemize}
\item \textbf{Consensus:} \textbf{18.5\%} is validated as the reliable answer.
\end{itemize}

\subsubsection{Role-Aligned Prompting}\label{role-aligned-prompting}

Assigns a specific persona (e.g., ``financial analyst'') to focus the model's knowledge and tone.

\textbf{Core Idea:} Constrain output by defining a professional context.

\textbf{Example:}

\begin{itemize}
\item \textbf{Prompt:} \texttt{"As a financial analyst, explain why a central bank raises interest rates."} → The role ensures a domain-specific, knowledge-grounded response.
\end{itemize}

Prompt engineering offers a \textbf{low-cost, agile} method to reduce hallucinations. By refining instructions, examples, and context, developers can significantly enhance factual accuracy \textbf{without model retraining}.

\subsection{Decoding Control}\label{decoding-control}

Decoding control mitigates hallucinations introduced by \textbf{inference-time sampling}. By constraining the token selection process, these techniques prioritize \textbf{factual consistency and stability over creative diversity} \cite{shuster2022blenderbot}.

\subsubsection{Sampling Parameters (Temperature, Top-k, Top-p)}\label{sampling-parameters-temperature-top-k-top-p}

These hyperparameters govern the randomness of token selection, acting as a dial between creativity and reliability.

\textbf{Core Idea:} Restrict the model's choice of next tokens to high-probability, predictable candidates.

\begin{itemize}
\item \textbf{Temperature:} Scales the logits before softmax. \textbf{Low T (e.g., 0.3)} sharpens the distribution, favoring almost-certain tokens. \textbf{High T (e.g., 1.2)} flattens it, allowing less likely tokens.
\item \textbf{Top-k:} Samples only from the \texttt{k} most probable next tokens.
\item \textbf{Top-p (Nucleus Sampling):} Samples only from tokens whose cumulative probability exceeds a threshold \texttt{p} (e.g., 0.9).
\end{itemize}

\textbf{Example:}

\begin{itemize}
\item \textbf{Prompt:} \texttt{"What is the capital of France?"}
\begin{itemize}
\item \textbf{High Temperature/Top-p:} \texttt{"Paris, though some consider Lyon a cultural capital."} (Creative, but potentially speculative)
\item \textbf{Low Temperature/Top-p:} \texttt{"Paris."} (Focused, factual)
\end{itemize}
\end{itemize}

\subsubsection{Self-Consistency Validation}\label{self-consistency-validation}

This technique runs the model multiple times with the same prompt and stochastic sampling, then aggregates the results to find a \textbf{consensus}, similar to what we discussed in Self-Consistency Prompting.

\textbf{Core Idea:} Identify the most reliable answer through \textbf{majority vote} across multiple generations.

\textbf{Implementation:} Use a \textbf{moderate temperature (\textgreater0)} to generate diverse reasoning paths, then select the most frequent final answer.

\textbf{Example:}

\begin{itemize}
\item \textbf{Prompt:} \texttt{"Who founded PayPal?"} (run 5 times)
\item \textbf{Outputs:} All list \texttt{"Elon Musk, Peter Thiel, and Max Levchin"} (in varying orders).
\item \textbf{Result:} High-confidence, consistent answer. Diverging answers would flag uncertainty.
\end{itemize}

\subsubsection{Constrained Decoding}\label{constrained-decoding}

Forces the model's output to adhere to a predefined structure or vocabulary, often drawn from an external knowledge source.

\textbf{Core Idea:} \textbf{Lexically constrain} generation to only include tokens or phrases from a verified set \cite{burns2023ccs}.

\textbf{Example:}

\begin{itemize}
\item \textbf{Task:} Generate a response about a company's credit rating using \textbf{only} the current ratings from a retrieved S\&P Global report.
\item \textbf{Prompt:} \texttt{"According to the latest report, Microsoft's credit rating is \_\_\_\_\_\_."}
\item \textbf{Constrained Output:} The model can only select from \texttt{\{AAA, AA+, AA\}}, preventing hallucination of incorrect names.
\end{itemize}

Decoding control acts as a \textbf{real-time reliability filter} during inference, ensuring generated text remains stable, consistent, and grounded---a critical safeguard in financial, legal, and analytical applications.

\subsection{Context Summarization for Length Management}\label{context-summarization-for-length-management}

When dealing with lengthy source documents that exceed model context windows or contain redundant information, \textbf{context summarization} serves as a powerful hallucination mitigation technique. This approach processes long contexts through intermediate summarization before final generation, addressing several hallucination triggers including attention dilution, context window limitations, and information overload \cite{galileo2024summarization,agenta2024context}.

\textbf{Core Idea: Map-Reduce Summarization} The most effective implementation follows a map-reduce pattern \cite{galileo2024summarization}:

\begin{enumerate}
\def\labelenumi{\arabic{enumi}.}
\item \textbf{Map Phase:} Divide the source document into coherent chunks and generate individual summaries for each segment
\item \textbf{Reduce Phase:} Synthesize the segment summaries into a final comprehensive summary
\item \textbf{Generation:} Use the condensed summary as context for the final task execution
\end{enumerate}

This approach is particularly valuable for financial document processing, where extracting key information from lengthy reports (e.g., annual filings, multi-page contracts) requires maintaining coherence across thousands of tokens \cite{galileo2024summarization}.

\textbf{Implementation Considerations}

\begin{itemize}
\item \textbf{Chunking Strategy:} Use semantic chunking at topic boundaries rather than arbitrary token counts to preserve contextual integrity
\item \textbf{Hierarchical Processing:} For extremely long documents, employ multiple summarization layers to gradually distill information while preserving critical insights
\item \textbf{Overlap Management:} Implement 10-20\% chunk overlap to ensure continuous topics aren't artificially split
\end{itemize}

\textbf{Example:} A 150-page annual report undergoes initial segmentation into 10 thematic sections. Each section is summarized to capture key financial metrics and risk disclosures. These section summaries are then synthesized into a comprehensive 2-page executive summary, which serves as the context for extracting specific data points like debt-to-equity ratios or liquidity metrics, significantly reducing hallucination risks from context overload.

\subsection{Fine-Tuning}\label{fine-tuning}

When hallucinations stem from \textbf{fundamental gaps in the model's pre-training}, fine-tuning provides a targeted solution. This process involves additional training cycles to systematically correct model behavior, making it more factual and reliable.

\subsubsection{Adversarial Training}\label{adversarial-training}

This technique improves robustness by training the model on deliberately challenging or misleading inputs.

\textbf{Core Idea:} Fortify the model against edge cases and \textbf{deceptive queries} by exposing it to \texttt{"trick questions."}

\textbf{Method:} Generate or curate adversarial examples where the model is prone to hallucinate and include the correct responses in the training data.

\textbf{Example:}

\begin{itemize}
\item \textbf{Adversarial Prompt:} \texttt{"Summarize the 2025 annual report for Bank of X."}
\item \textbf{Before Training:} Model fabricates a plausible-sounding summary.
\item \textbf{After Training:} Model responds, \texttt{"I cannot find a 2025 report for the Bank of X, as it is a fictional institution."}
\end{itemize}

\subsubsection{Contrastive Learning}\label{contrastive-learning}

This method teaches the model to distinguish correct from incorrect information by learning from paired examples.

\textbf{Core Idea:} Shape the model's internal representations by contrasting \textbf{factual outputs} against \textbf{hallucinated or incorrect counterparts} \cite{jiang2024hacl}.

\textbf{Method:} During training, the model is presented with triplets: an input, a positive (factual) completion, and a negative (hallucinated) completion. It learns to maximize the similarity to the positive example and minimize it to the negative.

\textbf{Example:}

\begin{itemize}
\item \textbf{Input:} \texttt{"The primary policy tool of the Bank of Canada is the..."}
\item \textbf{Positive Completion:} \texttt{"...overnight lending rate."}
\item \textbf{Negative Completion:} \texttt{"...exchange rate with the US dollar."}
\item \textbf{Result:} The model learns to strongly associate the input with the correct factual completion.
\end{itemize}

\subsubsection{Instruction Tuning \& RLHF}\label{instruction-tuning-rlhf}

These techniques align the model with human preferences, prioritizing attributes like factuality, helpfulness, and harmlessness.

\textbf{Core Idea:} Use human feedback as a training signal to steer the model away from generating unverified or incorrect content \cite{openai2023gpt4}.

\begin{itemize}
\item \textbf{Instruction Tuning:} Supervised training on a dataset of (instruction, desired response) pairs that demonstrate factual, verifiable answers.
\item \textbf{Reinforcement Learning from Human Feedback (RLHF):} A reward model is trained to score outputs based on human preferences, which is then used to fine-tune the main model via reinforcement learning.
\end{itemize}

\textbf{Example:}

\begin{itemize}
\item \textbf{Task:} Extract the ``Effective Date'' from a loan agreement.
\item \textbf{Document Text:} \texttt{"...this Credit Agreement is entered into as of March 15, 2024 (the 'Effective Date')..."}
\item \textbf{Before Fine-Tuning:} Model hallucinates a date not present in the document, likely by associating ``agreement'' with a signing date from its training data.
\begin{itemize}
\item \textbf{Model Output:} \texttt{"January 1, 2024"} \ding{55} (Critical hallucination)
\end{itemize}
\item \textbf{After Instruction Tuning/RLHF:} The model has been tuned on examples that reward strict adherence to the provided text and penalize fabrication. It learns to output ``Not Stated'' if the field is ambiguous or missing, rather than guessing.
\begin{itemize}
\item \textbf{Model Output:} \texttt{"March 15, 2024"} \ding{51} (Correctly extracted from text)
\end{itemize}
\end{itemize}

Fine-tuning is \textbf{resource-intensive} but provides a \textbf{deep, systematic correction} to a model's knowledge and behavior, offering a more permanent solution to hallucination than inference-time mitigations. Recent work, however, highlights that the effectiveness of RLHF is sensitive to reward design and annotation quality, and may not consistently penalize subtle factual errors \cite{yue2025does}.

\subsection{Open vs. Closed Weight Models for Mitigation}\label{open-vs.-closed-weight-models-for-mitigation}

The choice between open and closed-weight models significantly impacts which mitigation strategies can be deployed and how effectively they address potential hallucination sources. Table~\ref{table3} compares the mitigation capabilities across these model types.

\begin{table}[ht!]
    \centering
    \small
    \caption{Model Accessibility and Hallucination Mitigation Capabilities}
    \label{table3}
    \rowcolors{2}{lighterredforrow}{lightredforrow}
    \setlength{\tabcolsep}{10pt}
    \renewcommand{\arraystretch}{1.4}
    \arrayrulecolor{bordercolor}
    \begin{tabularx}{0.95\textwidth}{>{\raggedright\arraybackslash}X>{\raggedright\arraybackslash}X>{\raggedright\arraybackslash}X}
    \specialrule{1pt}{0pt}{0pt}
    
    \rowcolor{headerred}
    \textbf{\textcolor{white}{Mitigation Strategy}} & 
    \textbf{\textcolor{white}{Open-Weight Models}} & 
    \textbf{\textcolor{white}{Closed-Weight Models}} \\
    \arrayrulecolor{white}\specialrule{2pt}{0pt}{0pt}\arrayrulecolor{bordercolor}
    
    \textcolor{tablefontcolor}{\textbf{External Knowledge Grounding}} & 
    \textcolor{tablefontcolor}{Full integration with RAG systems and custom knowledge bases} & 
    \textcolor{tablefontcolor}{Limited to provider-supported retrieval and grounding features} \\
    \arrayrulecolor{white}\specialrule{1pt}{0pt}{0pt}\arrayrulecolor{bordercolor}
    
    \textcolor{tablefontcolor}{\textbf{Confidence Calibration}} & 
    \textcolor{tablefontcolor}{Full control over calibration methods} & 
    \textcolor{tablefontcolor}{Limited to output-level adjustments} \\
    \arrayrulecolor{white}\specialrule{1pt}{0pt}{0pt}\arrayrulecolor{bordercolor}
    
    \textcolor{tablefontcolor}{\textbf{Prompt Engineering}} & 
    \textcolor{tablefontcolor}{Full flexibility in prompt design and optimization} & 
    \textcolor{tablefontcolor}{Available but constrained by provider's prompt processing} \\
    \arrayrulecolor{white}\specialrule{1pt}{0pt}{0pt}\arrayrulecolor{bordercolor}
    
    \textcolor{tablefontcolor}{\textbf{Decoding Control}} & 
    \textcolor{tablefontcolor}{Complete control over temperature, top-k, top-p, and constrained decoding} & 
    \textcolor{tablefontcolor}{Limited to provider-exposed decoding parameters} \\
    \arrayrulecolor{white}\specialrule{1pt}{0pt}{0pt}\arrayrulecolor{bordercolor}
    
    \textcolor{tablefontcolor}{\textbf{Context Summarization}} & 
    \textcolor{tablefontcolor}{More necessary due to smaller context windows and weaker long-context retention; requires tighter summarization.} & 
    \textcolor{tablefontcolor}{Less necessary because of very large context windows; used mainly for efficiency and reducing noise.} \\
    \arrayrulecolor{white}\specialrule{1pt}{0pt}{0pt}\arrayrulecolor{bordercolor}
    
    \textcolor{tablefontcolor}{\textbf{Fine-Tuning}} & 
    \textcolor{tablefontcolor}{Complete parameter control including custom RLHF, adversarial training, and contrastive learning} & 
    \textcolor{tablefontcolor}{Limited to provider-supported tuning} \\
    
    \specialrule{1pt}{0pt}{0pt}
    \end{tabularx}
\end{table}

Open-weight models offer \textbf{maximum flexibility} for implementing comprehensive mitigation strategies, particularly for addressing potential model-level issues through fine-tuning and architectural changes. However, their smaller context windows also make context summarization a critical necessity, e.g., in the face of lengthy documents. Closed-weight models, while operationally simpler, provide \textbf{limited intervention capabilities}, often restricting mitigation to prompt engineering and basic decoding controls. For these models, context summarization remains a valuable tool for managing extreme document length and reducing noise, even with their larger native context windows.

\subsection{The Inevitability of Hallucinations}\label{the-inevitability-of-hallucinations}

Despite sophisticated detection and mitigation strategies, \textbf{hallucinations cannot be completely eliminated from LLM and LRM systems}. The statistical nature of language modeling means these systems operate by \textbf{predicting plausible sequences} rather than verifying ground truth. Additionally, the dynamic nature of real-world knowledge creates inherent gaps between training data and current information. The framework presented here acknowledges these limitations while providing systematic approaches to \textbf{manage hallucination risks} rather than eliminate them entirely.

The mitigation strategies presented in this section form a comprehensive toolbox for addressing potential hallucination sources while acknowledging the inherent limitations of generative AI systems. By selecting and combining these techniques with awareness of their applicability to different potential root causes, financial institutions such as CIBC can systematically reduce hallucination risks while maintaining operational efficiency.

\section{Root Cause of Hallucination}\label{4-hallucination-source-hypothesis}

To be \textbf{root cause aware}, the framework must be guided by a structured understanding of \emph{why} hallucinations occur. This allows detection signals to be interpreted as indicators of probable failure categories and enables the selection of the most appropriate mitigation levers. Hallucinations are not random; they are systematic failures that can be traced to specific weaknesses in the model's knowledge, reasoning, or the context it operates within \cite{dahl2024legal,huang2025survey}.

We categorize the primary sources of hallucinations into three interconnected domains: issues arising from the \textbf{Model's Training and Design}, limitations inherent in the \textbf{Training Data}, and challenges presented by the \textbf{Context at Inference Time}. This taxonomy provides the conceptual map for root cause awareness, enabling the framework to link observable failure patterns to the most probable categories of underlying issues.

\subsection{Model Training \& Design}\label{model-training-design}

This category encompasses hallucinations stemming from the fundamental architecture and optimization process of the model itself. These are often the most challenging to address, as they are baked into the model's core operational principles.

\subsubsection{Architectural Constraints}\label{architectural-constraints}

The transformer architecture, while powerful, has inherent limitations. A finite \textbf{context window} can prevent the model from maintaining coherence over long documents, leading to intrinsic contradictions \cite{dahl2024legal}. Furthermore, the autoregressive nature of next-token prediction prioritizes local fluency over global factual consistency, a phenomenon often described as the model ``compulsively completing the pattern'' even if it requires fabrication \cite{ji2023survey}. For complex reasoning tasks, standard architectures may lack the necessary depth or specialized modules to perform multi-step logic reliably, resulting in semantic hallucinations where the conclusion does not follow from the premises \cite{wang2025race}.

\subsubsection{Training Objectives and Fine-Tuning}\label{training-objectives-and-fine-tuning}

The standard pre-training objective of maximizing the likelihood of the next token does not explicitly reward factual accuracy. This can lead models to generate plausible-sounding but incorrect sequences that are statistically likely in the training data \cite{openaiblog2023}. Techniques like \textbf{Reinforcement Learning from Human Feedback (RLHF)} can inadvertently exacerbate this by over-optimizing for styles of response that humans prefer (e.g., confident, fluent) at the expense of factuality, a known issue called ``reward hacking'' \cite{gao2023harms}. Similarly, domain-specific fine-tuning on small, imperfect datasets can cause the model to overfit to spurious correlations present in that data \cite{brown2020language}.

\subsubsection{Decoding Dynamics}\label{decoding-dynamics}

The strategy used to select tokens during generation (inference) can introduce variability and error. Techniques like \textbf{nucleus sampling (top-p)} or high \textbf{temperature} settings, which increase creativity and diversity, also heighten the risk of the model veering into nonsensical or unfactual territory \cite{holtzman2020curious}. Conversely, overly greedy strategies like beam search can lead to repetitive and degenerate outputs, another form of hallucination \cite{welleck2019neural}.

\subsection{Training Data Induced}\label{training-data-induced}

LLMs are knowledge repositories distilled from their training corpora. Imperfections, biases, and limitations in this data are directly reflected in the model's outputs.

\subsubsection{Knowledge Gaps and Outdated Information}\label{knowledge-gaps-and-outdated-information}

A model can only be as current and comprehensive as its training data. \textbf{Hallucinations of omission} occur when the model lacks specific knowledge and, instead of acknowledging ignorance, confabulates an answer based on vague semantic associations \cite{ji2023survey}. This is particularly acute for information with a rapid update cycle, such as financial regulations, corporate earnings, or current events. A model trained on data with a 2023 cutoff will confidently hallucinate answers about 2025 events, posing a significant risk in time-sensitive domains \cite{magesh2025legalrag}.

\subsubsection{Noisy, Biased, or Spurious Correlations}\label{noisy-biased-or-spurious-correlations}

Large-scale web-crawled datasets are rife with inaccuracies, contradictions, and biased viewpoints. The model may learn and reproduce these falsehoods or biases as if they were facts \cite{bender2021dangers}. Furthermore, it can learn \textbf{spurious correlations}---for example, associating a company's name with an outdated stock ticker or a CEO from a previous decade. When generating text, these incorrect associations manifest as extrinsic factual hallucinations \cite{mccoy2019right}.

\subsubsection{Source-Reference Divergence}\label{source-reference-divergence}

In tasks like summarization, models are often trained on datasets where the human-written ``reference'' summary contains information not explicitly stated in the source document. The model learns that summarization involves a degree of abstraction and invention, which can lead to unsupported claims when applied in high-stakes, fact-bound scenarios \cite{pagnoni2021understanding}.

\subsection{Context at Inference Time}\label{context-at-inference-time}

Finally, hallucinations can be triggered or amplified by the specific context provided to the model during use. This highlights that the fault is not always with the model itself, but with how it is prompted and deployed.

\subsubsection{Ambiguous or Underspecified Prompts}\label{ambiguous-or-underspecified-prompts}

A prompt that lacks necessary detail or context forces the model to ``fill in the blanks.'' For instance, asking an LLM to ``summarize the company's financial performance'' without providing the specific report invites the model to generate a generic, potentially fabricated summary based on its parametric knowledge of typical financial reports, which may not match the actual company in question \cite{dahl2024legal}. This is a common source of domain-specific hallucinations.

\subsubsection{Contradictory or Misleading Context}\label{contradictory-or-misleading-context}

In Retrieval-Augmented Generation (RAG) systems, if the retrieved documents are irrelevant or contain conflicting information, the model may struggle to synthesize a coherent answer, leading to inconsistencies or fabrications as it attempts to reconcile the noise \cite{lewis2020rag}. The model's tendency to please the user by following the immediate context can override its internal knowledge, a phenomenon known as \textbf{contextual override} \cite{liu2021context}.

\subsubsection{Domain and Task Shift}\label{domain-and-task-shift}

When a model is applied to a domain or task significantly different from its training distribution, its performance can degrade unpredictably. A general-purpose LLM prompted to perform a complex financial risk calculation without specific fine-tuning may produce a semantically coherent but mathematically flawed reasoning chain, resulting in a critical reasoning hallucination \cite{wang2025race}.

\noindent
This taxonomy is not employed to definitively attribute a single root cause to each error. Instead, its purpose is to instill a state of \textbf{root cause awareness} within the framework. Detection signals are interpreted as indicators that point toward one or more probable categories of underlying issues.

This awareness directly and efficiently informs the selection of mitigation strategies. Specific failure patterns trigger targeted interventions from the appropriate mitigation toolbox. This targeted response, guided by a probabilistic understanding of potential failure sources, forms the core of our tier-based hallucination management system discussed in Section~\ref{6-use-case-application} through a data extraction use case.

\section{Operational Framework \& Improvement Cycle}\label{5-operational-framework--continuous-improvement}

The proposed \textbf{Continuous Improvement Cycle} establishes a closed-loop operational framework that incrementally enhances model reliability through \emph{root cause--aware} detection and mitigation. Rather than presuming fixed sources of hallucination, the cycle learns from evidence surfaced during operation---continuously transforming detection outcomes into informed mitigation and refinement decisions.

This iterative process functions through four main stages: \textbf{Detection}, \textbf{Mitigation}, \textbf{Validation}, and \textbf{Refinement}. Each stage feeds the next, ensuring that the system evolves based on measurable outcomes rather than static assumptions.

\subsection{Detection-Initiated Evolution}\label{detection-initiated-evolution}

The cycle begins with \textbf{Detection}, where outputs are automatically screened for uncertainty, inconsistency, or factual deviation. Signals such as elevated entropy, fact-checking failures, or reasoning incoherence act as \emph{indicators} of potential hallucinations. Rather than directly attributing fault to a single root cause, these signals guide the system to the most relevant \emph{detection tier}---for example, whether the issue appears model-, context-, or data-related (see Section~\ref{6-use-case-application}). This evidence-driven awareness enables the framework to decide the most appropriate next diagnostic or mitigation step.

\subsection{Root Cause--Aware Mitigation}\label{root-causeaware-mitigation}

Once a hallucination is confirmed, \textbf{Mitigation} applies the most suitable intervention \emph{based on detected patterns}, not fixed rules. If the detection signal suggests uncertainty, the framework may apply \textbf{confidence calibration} or \textbf{decoding control}. If it reveals contextual inconsistency, \textbf{prompt optimization} or \textbf{intrinsic consistency checks} are triggered. If the evidence points toward outdated or incomplete information, \textbf{external grounding} or \textbf{fine-tuning} may follow. This selective approach ensures that interventions are proportionate, minimally invasive, and tailored to operational conditions.

\subsection{Validation and Feedback}\label{validation-and-feedback}

Following mitigation, outputs are re-evaluated using the same detection metrics that initiated the process. If hallucinations persist, the framework records them as \emph{residual errors} and escalates the case to the next appropriate detection tier or to human review. This \textbf{Validation Feedback} ensures every corrective step is data-validated and that progress is measurable across cycles.

\subsection{Adaptive Refinement}\label{adaptive-refinement}

Over successive iterations, the framework accumulates knowledge about which detection signals correspond to which mitigation strategies and under what contexts they succeed. This evolving understanding builds a dynamic \emph{root cause awareness}---a mapping between hallucination patterns, interventions, and outcomes. It enables the system to allocate computational or human oversight resources more efficiently and refine future detection and mitigation behavior.

In summary, the Continuous Improvement Cycle functions as a self-correcting operational backbone: detection identifies potential instability, mitigation applies the most context-appropriate correction, validation measures impact, and refinement captures learning. In Section~\ref{6-use-case-application}, this process is operationalized through a \textbf{tier-based detection and mitigation system}, where model-, context-, and data-aware mechanisms interact within this continuous improvement loop.

\section{Case Study - Data Extraction}\label{6-use-case-application}

To demonstrate the practical applicability of the proposed framework, we apply the \textbf{Operational Framework \& Improvement Cycle} to a representative use case: \textbf{data extraction} within a financial organization. This task involves automatically identifying, extracting, and structuring information from unstructured financial documents such as loan applications, contracts, or statements. Because of the high factual and regulatory sensitivity of such data, hallucination management becomes critical to ensure that extracted content is both \emph{accurate} and \emph{contextually faithful} to the source.

\subsection{Overview of the Tiered Detection and Mitigation Architecture}\label{overview-of-the-tiered-detection-and-mitigation-architecture}

\begin{figure}[ht]
    \centering
    \includegraphics[width=1\linewidth]{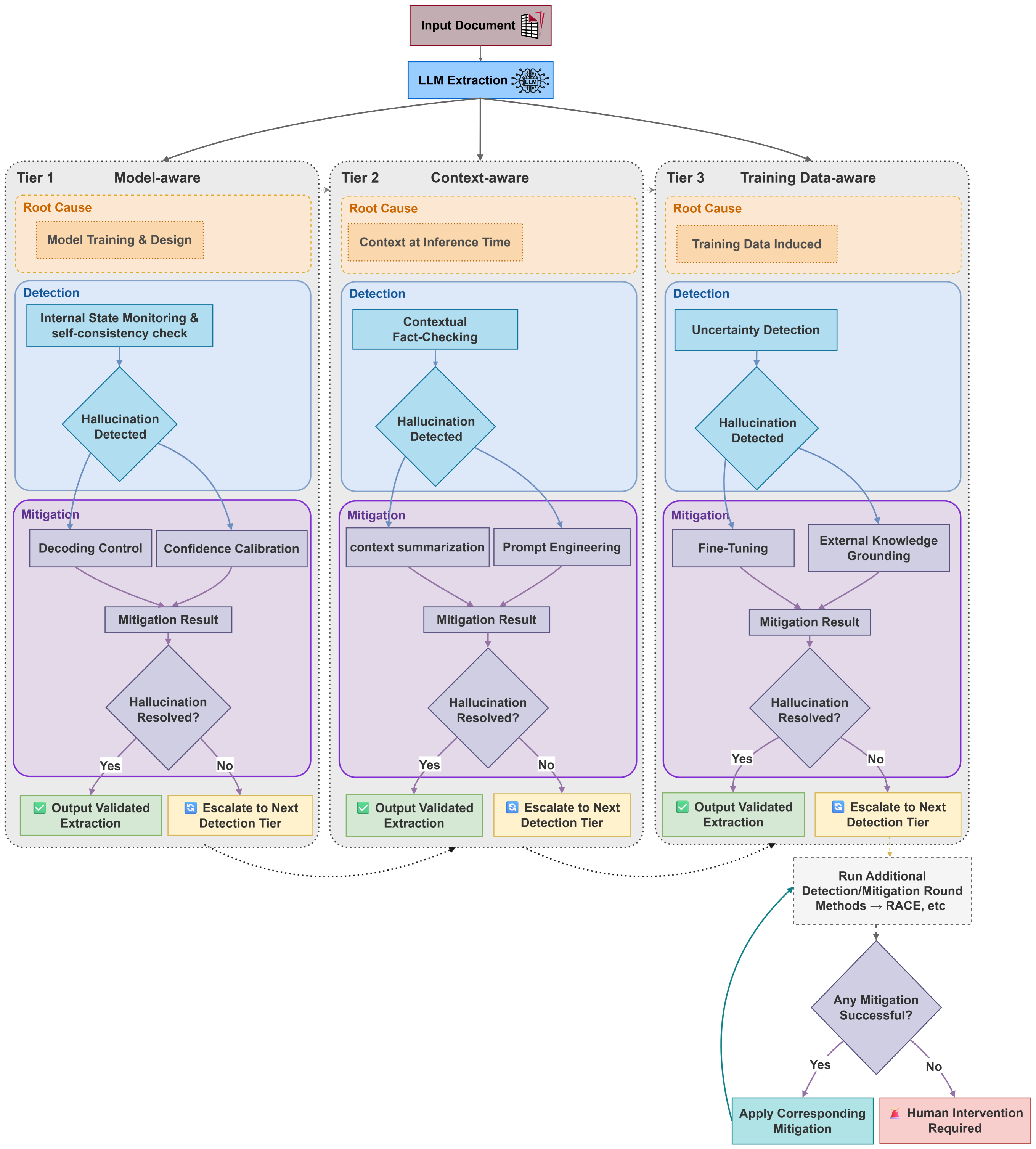}
    \caption{Tier-based hallucination detection and mitigation pipeline for the data extraction use case.}
    \label{fig3}
\end{figure}

The complete operational structure for this use case is illustrated in Figure~\ref{fig3}. It represents a \textbf{tier-based hallucination management system} integrated within the continuous improvement loop introduced in Section~\ref{5-operational-framework--continuous-improvement}. The framework is organized into three complementary tiers---\textbf{Model}, \textbf{Context}, and \textbf{Data}---each responsible for distinct yet interconnected sources of potential hallucination. Together, these tiers create a closed feedback loop in which detection, mitigation, and validation operate continuously to improve reliability over time.

\subsection{Model Tier -- Intrinsic Reliability Control}\label{model-tier-intrinsic-reliability-control}

The \textbf{Model Tier} focuses on hallucinations arising from the LLM's intrinsic behavior---uncalibrated confidence, excessive decoding randomness, or overgeneralized language generation. In the data extraction scenario, such hallucinations may appear as incorrectly generated entities (e.g., misidentified borrower names or fabricated monetary values). Detection methods include \textbf{temperature-scaled confidence estimation}, \textbf{logit entropy analysis}, and \textbf{self-consistency validation} across multiple generations. Upon detection, \textbf{temperature scaling}, \textbf{logit adjustment}, or \textbf{ensemble agreement filtering} are applied as mitigation mechanisms. Validated corrections from this tier are fed into the refinement stage of the continuous improvement cycle to update the model's calibration parameters.

\subsection{Context Tier -- Prompt and Instructional Grounding}\label{context-tier-prompt-and-instructional-grounding}

The \textbf{Context Tier} addresses hallucinations originating from \emph{instructional or situational misalignment} between the LLM and the data extraction task. For instance, the model might produce plausible but irrelevant outputs if the extraction prompt is ambiguous or lacks sufficient context about document layout or terminology. Detection at this tier relies on \textbf{semantic similarity checks} between generated fields and reference descriptions, as well as \textbf{reasoning-trace consistency scoring}. Mitigation occurs through \textbf{prompt optimization}, \textbf{instruction reweighting}, and/or \textbf{context summarization}, ensuring that the model remains properly grounded in the intended extraction context. This tier also interfaces directly with user or system-level feedback loops to refine prompting templates in subsequent cycles.

\subsection{Training Data Tier -- External Verification and Knowledge Alignment}\label{training-data-tier-external-verification-and-knowledge-alignment}

The \textbf{Training Data Tier} governs hallucinations stemming from incomplete or outdated external information. In the context of data extraction, these errors often manifest when the model produces values not verifiable within the document corpus---for example, referencing non-existent client identifiers or outdated financial terms. Detection involves \textbf{cross-source factual consistency checking} and \textbf{entity-level validation} against structured databases or OCR-derived metadata. When discrepancies are found, \textbf{retrieval-augmented grounding} or \textbf{fine-tuning with verified samples} are triggered. This ensures that the model's knowledge remains aligned with authentic, up-to-date data sources.

\subsection{Cross-Tier Feedback Integration}\label{cross-tier-feedback-integration}

Each of the three tiers operates semi-independently but shares \emph{feedback signals} through the continuous improvement loop (see Section~\ref{5-operational-framework--continuous-improvement}). For example, residual errors detected at the Data Tier may reveal previously unrecognized contextual weaknesses, prompting the Context Tier to refine its prompts. Similarly, recurring low-confidence detections at the Model Tier can inform re-training or calibration strategies that enhance performance across all tiers. This \textbf{cross-tier feedback} transforms the architecture from a linear correction pipeline into an adaptive, self-improving system.

\subsection{Operational Outcome}\label{operational-outcome}

Applying this architecture to data extraction yields several operational benefits:

\begin{enumerate}
\def\labelenumi{\arabic{enumi}.}
\item \textbf{Reduced hallucination frequency} through tier-targeted detection and calibration.
\item \textbf{Improved factual accuracy} in extracted entities and key-value mappings.
\item \textbf{Increased interpretability} by localizing errors to a specific tier and intervention type.
\item \textbf{Continuous performance growth} driven by data-backed refinement across detection and mitigation stages.
\end{enumerate}

In essence, the Data Extraction case study demonstrates how the proposed tier-based hallucination management framework (Figure~\ref{fig1}) can be instantiated for a high-stakes, text-intensive task. By embedding the detection--mitigation--validation cycle within three hierarchical tiers---Model, Context, and Data---the system ensures that hallucination control becomes both explainable and continuously improving. This approach generalizes naturally to other generative AI tasks where factual precision and interpretability are important.

\section{Conclusion}\label{7-conclusion}

This report has introduced a comprehensive, operational framework for the systematic management of hallucinations in Large Language Models (LLMs) and Large Reasoning Models (LRMs). Moving beyond one-size-fits-all solutions, the framework is built on a \textbf{continuous improvement cycle} that integrates \textbf{root cause-aware} detection and mitigation. This approach acknowledges that while hallucinations are an inherent byproduct of the statistical nature of generative AI, their risks can be systematically managed and reduced through a structured, diagnostic process.

\subsection{Key Contributions}\label{key-contributions}

The core contribution of this work is a \textbf{structured methodology} that transforms hallucination management from a reactive challenge into a proactive, continuous process. The framework's key components work in concert to achieve this:

\begin{itemize}
\item \textbf{Root Cause Awareness:} Grounded in a detailed taxonomy of hallucination sources---\textbf{Model, Data, and Context}---the framework ensures that detection signals are interpreted as indicators of potential underlying issues. This diagnostic awareness prevents the misapplication of mitigation strategies, enabling efficient and targeted interventions.
\item \textbf{A Unified Detection Toolkit:} We have detailed a multi-faceted suite of detection methods, from probabilistic and semantic entropy to advanced uncertainty estimation and the RACE framework. This allows for a nuanced understanding of different failure modes, whether they stem from token-level uncertainty, semantic inconsistency, or flawed reasoning.
\item \textbf{A Targeted Mitigation Toolbox:} The framework provides a stratified set of mitigation strategies---including external knowledge grounding, confidence calibration, prompt engineering, and fine-tuning---each designed to address specific categories of potential root causes identified during detection.
\item \textbf{An Operational, Tiered Architecture:} The \textbf{Data Extraction case study} demonstrates how the framework is instantiated in practice through a \textbf{tier-based system} (Model, Context, Data). This architecture operationalizes the continuous improvement cycle, creating a closed-loop system where detection, mitigation, validation, and refinement feed into one another, driving incremental gains in reliability.
\end{itemize}

\subsection{Practical Implications}\label{practical-implications}

For \textbf{AI practitioners and system architects}, this framework provides a blueprint for building more reliable and verifiable AI systems. It offers clear guidance on selecting detection metrics based on operational constraints (e.g., open vs.~closed-weight models, ground truth availability) and deploying targeted mitigations that are both effective and resource-efficient.

For \textbf{decision-makers and risk officers} in regulated industries like finance and legal services, the framework delivers a critical assurance: hallucination risks are not merely acknowledged but are actively managed through a verifiable, systematic process. It provides the transparency and structured controls necessary for responsible AI deployment, helping to mitigate financial, regulatory, and reputational risks.

\subsection{The Inevitability of Hallucinations and the Path Forward}\label{the-inevitability-of-hallucinations-and-the-path-forward}

It is crucial to recognize that hallucinations cannot be entirely eliminated. They are a fundamental trait of generative models that operate by predicting plausible sequences rather than accessing ground truth. Our framework accepts this reality and focuses on building resilient systems that can \textbf{identify, contain, and learn from} these failures, rather than pursuing the unattainable goal of perfect accuracy.

By adopting the root cause-aware, continuous improvement framework outlined in this paper, organizations can harness the transformative power of LLMs and LRMs with greater confidence, building generative AI systems that are not only powerful but also progressively more trustworthy and reliable.

\section*{Acknowledgments}

The authors would like to extend sincere thanks to colleagues at CIBC for their valuable insights and constructive feedback during the preparation of this work. In particular, we are grateful to Gareth Stoyle for his thoughtful advice, which has strengthened this paper.

\bibliographystyle{unsrt}
\bibliography{references}

@article{bender2021dangers,
  title        = {On the Dangers of Stochastic Parrots: Can Language Models Be Too Big?},
  author       = {Bender, Emily M. and Gebru, Timnit and McMillan-Major, Angelina and Shmitchell, Shmargaret},
  journal      = {Proceedings of the ACM Conference on Fairness, Accountability, and Transparency},
  year         = {2021},
  pages        = {610--623}
}

@article{brown2020language,
  title        = {Language Models are Few-Shot Learners},
  author       = {Brown, Tom B. and Mann, Benjamin and Ryder, Nick and others},
  journal      = {Advances in Neural Information Processing Systems},
  volume       = {33},
  pages        = {1877--1901},
  year         = {2020}
}

@inproceedings{burns2023ccs,
  title        = {Discovering Hallucinations in Large Language Models via Contrast-Consistent Search},
  author       = {Burns, Collin and Chen, Haotian and et al.},
  booktitle    = {Advances in Neural Information Processing Systems (NeurIPS)},
  year         = {2023},
  url          = {https://arxiv.org/abs/2303.08896}
}

@article{dahl2024legal,
  title        = {Large Legal Fictions: Profiling Legal Hallucinations in Large Language Models},
  author       = {Dahl, Matthew and Magesh, Varun and Suzgun, Mirac and Ho, Daniel E.},
  journal      = {arXiv preprint arXiv:2401.01301},
  year         = {2024},
  note         = {Documentation of hallucination rates across LLMs on legal cases; intrinsic vs extrinsic errors}
}

@article{farquhar2024semantic,
  title        = {Semantic Uncertainty: Measuring Meaning-Level Confidence for Hallucination Detection},
  author       = {Farquhar, Sebastian and Marjieh, R. and Gal, Y.},
  journal      = {Nature},
  volume       = {627},
  pages        = {58--65},
  year         = {2024},
  doi          = {10.1038/s41586-024-07421-0}
}

@article{gao2023harms,
  title        = {Harms from Increasingly Agentic AI Systems},
  author       = {Gao, Leo and Schulman, John and Hilton, Jacob},
  journal      = {arXiv preprint arXiv:2302.10329},
  year         = {2023}
}

@article{holtzman2020curious,
  title        = {The Curious Case of Neural Text Degeneration},
  author       = {Holtzman, Ari and Buys, Jan and Du, Maxwell and Forbes, Maxwell and Choi, Yejin},
  journal      = {International Conference on Learning Representations},
  year         = {2020}
}

@article{jiang2024hacl,
  title        = {Hallucination-Augmented Contrastive Learning for Multimodal Large Language Models},
  author       = {Jiang, Y. and Park, S. and Lee, J.},
  journal      = {arXiv preprint},
  year         = {2024},
  url          = {https://arxiv.org/abs/2312.06968}
}

@article{ji2023survey,
  title        = {Survey of Hallucination in Natural Language Generation},
  author       = {Ji, Ziwei and Lee, Nayeon and Frieske, Rita and Yu, Tiezheng and Su, Dan and Xu, Yan and Ishii, Etsuko and Bang, Yejin and Dai, Wenliang and Madotto, Andrea and others},
  journal      = {ACM Computing Surveys},
  volume       = {55},
  number       = {12},
  pages        = {1--38},
  year         = {2023}
}

@article{kang2023finance,
  title        = {Deficiency of Large Language Models in Finance: An Empirical Examination of Hallucination},
  author       = {Kang, Haoqiang and Liu, Xiao‐Yang},
  journal      = {arXiv preprint arXiv:2311.15548},
  year         = {2023},
  note         = {Empirical study of hallucination behavior of LLMs on financial tasks}
}

@inproceedings{kojima2022cot,
  title        = {Large Language Models are Zero-Shot Reasoners},
  author       = {Kojima, Takeshi and Gu, Shixiang Shane and Reid, Jonathan and Matsuo, Yutaka and Iwasawa, Yusuke},
  booktitle    = {NeurIPS 2022},
  year         = {2022},
  url          = {https://arxiv.org/abs/2205.11916}
}

@article{xie2024models,
  title={What Are Models Thinking About? Understanding Large Language Model Hallucinations Through Model Internal State Analysis},
  author={Xie, Yuxi and Wang, Zhen and Li, Jie and Zhang, Yanzhe and Zhang, Rui and Wang, Di},
  journal={arXiv preprint arXiv:2406.18308},
  year={2024}
}

@article{huang2025survey,
  title={A survey on hallucination in large language models: Principles, taxonomy, challenges, and open questions},
  author={Huang, Lei and Yu, Weijiang and Ma, Weitao and Zhong, Weihong and Feng, Zhangyin and Wang, Haotian and Chen, Qianglong and Peng, Weihua and Feng, Xiaocheng and Qin, Bing and others},
  journal={ACM Transactions on Information Systems},
  volume={43},
  number={2},
  pages={1--55},
  year={2025},
  publisher={ACM New York, NY}
}

@inproceedings{guo2017calibration,
  title={On calibration of modern neural networks},
  author={Guo, Chuan and Pleiss, Geoff and Sun, Yu and Weinberger, Kilian Q},
  booktitle={International conference on machine learning},
  pages={1321--1330},
  year={2017},
  organization={PMLR}
}

@article{shorinwa2025survey,
  title={A survey on uncertainty quantification of large language models: Taxonomy, open research challenges, and future directions},
  author={Shorinwa, Ola and Mei, Zhiting and Lidard, Justin and Ren, Allen Z and Majumdar, Anirudha},
  journal={ACM Computing Surveys},
  year={2025},
  publisher={ACM New York, NY}
}

@article{kadavath2022language,
  title={Language models (mostly) know what they know},
  author={Kadavath, Saurav and Conerly, Tom and Askell, Amanda and Henighan, Tom and Drain, Dawn and Perez, Ethan and Schiefer, Nicholas and Hatfield-Dodds, Zac and DasSarma, Nova and Tran-Johnson, Eli and others},
  journal={arXiv preprint arXiv:2207.05221},
  year={2022}
}

@inproceedings{lewis2020rag,
  title        = {Retrieval-Augmented Generation for Knowledge-Intensive NLP Tasks},
  author       = {Lewis, Patrick and Perez, Ethan and Piktus, Aleksandra and Petroni, Fabio and others},
  booktitle    = {Advances in Neural Information Processing Systems (NeurIPS)},
  year         = {2020},
  url          = {https://arxiv.org/abs/2005.11401}
}

@article{liu2021context,
  title        = {Context-aware Language Modeling for Goal-oriented Dialogue Systems},
  author       = {Liu, Qian and others},
  journal      = {arXiv preprint arXiv:2105.13855},
  year         = {2021}
}

@article{magesh2025legalRAG,
  title        = {Hallucination-Free? Assessing the Reliability of Leading AI Legal Research Tools},
  author       = {Magesh, Varun and Surani, Faiz and Dahl, Matthew and Suzgun, Mirac and Manning, Christopher D. and Ho, Daniel E.},
  journal      = {Journal of Empirical Legal Studies},
  volume       = {0},
  number       = {1},
  pages        = {1–27},
  year         = {2025},
  doi          = {10.1111/jels.12413},
  note         = {Empirical evaluation of Lexis+ AI, Westlaw AI, etc., showing nontrivial hallucination rates even for RAG-based legal tools}
}

@article{mccoy2019right,
  title        = {Right for the Wrong Reasons: Diagnosing Model Syntactic Heuristics in Natural Language Inference},
  author       = {McCoy, Tom and Pavlick, Ellie and Linzen, Tal},
  journal      = {Proceedings of the 57th Annual Meeting of the Association for Computational Linguistics},
  pages        = {3428--3448},
  year         = {2019}
}

@misc{openai2023gpt4,
  title        = {GPT-4 Technical Report},
  author       = {OpenAI},
  year         = {2023},
  url          = {https://arxiv.org/abs/2303.08774}
}

@misc{openaiblog2023,
  title        = {How should AI systems behave, and who should decide?},
  author       = {OpenAI},
  year         = {2023},
  howpublished = {\url{https://openai.com/index/how-should-ai-systems-behave/}},
  note         = {Accessed: 2025-03-15}
}

@article{pagnoni2021understanding,
  title        = {Understanding Factuality in Abstractive Summarization with FRANK: A Benchmark for Factuality Metrics},
  author       = {Pagnoni, Artidoro and Balachandran, Vidhisha and Tsvetkov, Yulia},
  journal      = {Proceedings of the 2021 Conference of the North American Chapter of the Association for Computational Linguistics: Human Language Technologies},
  pages        = {4812--4829},
  year         = {2021}
}

@inproceedings{shuster2022blenderbot,
  title        = {BlenderBot 3: a deployed conversational agent that continually learns to responsibly engage},
  author       = {Shuster, Kurt and Xu, Jing and Komeili, Mojtaba and others},
  booktitle    = {arXiv preprint arXiv:2208.03188},
  year         = {2022},
  url          = {https://arxiv.org/abs/2208.03188}
}

@inproceedings{wang2023selfconsistency,
  title        = {Self-Consistency Improves Chain-of-Thought Reasoning in Language Models},
  author       = {Wang, Xuezhi and Wei, Jason and Schuurmans, Dale and others},
  booktitle    = {International Conference on Learning Representations (ICLR)},
  year         = {2023},
  url          = {https://arxiv.org/abs/2203.11171}
}

@article{wang2025race,
  title        = {RACE: Reasoning and Answer Consistency Evaluation for Large Reasoning Models},
  author       = {Wang, T. and Mitchell, E. and Azaria, A.},
  journal      = {arXiv preprint},
  year         = {2025},
  url          = {https://arxiv.org/abs/2506.04832}
}

@article{welleck2019neural,
  title        = {Neural Text Generation with Unlikelihood Training},
  author       = {Welleck, Sean and Kulikov, Ilia and Roller, Stephen and Dinan, Emily and Cho, Kyunghyun and Weston, Jason},
  journal      = {International Conference on Learning Representations},
  year         = {2020}
}

@article{zhang2025faith,
  title        = {FAITH: A Framework for Assessing Intrinsic Tabular Hallucinations in Finance},
  author       = {Zhang, Mengao and Fu, Jiayu and Warrier, Tanya and Wang, Yuwen and Tan, Tianhui and Huang, Ke‐wei},
  journal      = {arXiv preprint arXiv:2508.05201},
  year         = {2025},
  note         = {Evaluates intrinsic hallucinations for tabular financial data, S\&P 500 annual reports, etc.}
}

@article{gal2016dropout,
  title={Dropout as a Bayesian approximation: Representing model uncertainty in deep learning},
  author={Gal, Yarin and Ghahramani, Zoubin},
  journal={Proceedings of the 33rd International Conference on Machine Learning (ICML)},
  year={2016}
}

@article{kendall2017uncertainties,
  title={What Uncertainties Do We Need in Bayesian Deep Learning for Computer Vision?},
  author={Kendall, Alex and Gal, Yarin},
  journal={Advances in Neural Information Processing Systems (NeurIPS)},
  year={2017}
}

@inproceedings{lakshminarayanan2017simple,
  title={Simple and Scalable Predictive Uncertainty Estimation using Deep Ensembles},
  author={Lakshminarayanan, Balaji and Pritzel, Alexander and Blundell, Charles},
  booktitle={Advances in Neural Information Processing Systems (NeurIPS)},
  year={2017}
}

@misc{pleiss_blog_calibration,
  title   = {Calibrating Neural Networks},
  author  = {Pleiss, Geoff},
  year    = {2019},
  note    = {Blog post},
  url     = {https://geoffpleiss.com/blog/nn_calibration.html}
}

@article{adaptive_temp_scaling2024,
  title     = {Adaptive Temperature Scaling for Robust Calibration of Deep Neural Networks},
  author    = {Hu, Xin and Zhang, Zhenyu and Wang, Tao and Luo, Yuan},
  journal   = {Neural Computing and Applications},
  year      = {2024},
  doi       = {10.1007/s00521-024-09505-4},
  publisher = {Springer}
}

@inproceedings{zadrozny2002transforming,
  title     = {Transforming Classifier Scores into Accurate Multiclass Probability Estimates},
  author    = {Zadrozny, Bianca and Elkan, Charles},
  booktitle = {Proceedings of the Eighth ACM SIGKDD International Conference on Knowledge Discovery and Data Mining (KDD)},
  pages     = {694--699},
  year      = {2002},
  publisher = {ACM}
}

@inproceedings{niculescu2005predicting,
  title     = {Predicting Good Probabilities with Supervised Learning},
  author    = {Niculescu-Mizil, Alexandru and Caruana, Rich},
  booktitle = {Proceedings of the 22nd International Conference on Machine Learning (ICML)},
  pages     = {625--632},
  year      = {2005},
  publisher = {ACM}
}

@inproceedings{vadera2021posthoc,
  title     = {Post-hoc Loss-Calibration for Bayesian Neural Networks},
  author    = {Vadera, Meet P. and Ghosh, Soumya and Ng, Kenney and Marlin, Benjamin M.},
  booktitle = {Proceedings of the 37th Conference on Uncertainty in Artificial Intelligence (UAI)},
  pages     = {338--348},
  year      = {2021},
  url       = {https://proceedings.mlr.press/v161/vadera21a/vadera21a.pdf}
}

@inproceedings{blundell2015weight,
  title     = {Weight Uncertainty in Neural Networks},
  author    = {Blundell, Charles and Cornebise, Julien and Kavukcuoglu, Koray and Wierstra, Daan},
  booktitle = {International Conference on Machine Learning (ICML)},
  pages     = {1613--1622},
  year      = {2015}
}

@book{neal2012bayesian,
  title     = {Bayesian Learning for Neural Networks},
  author    = {Neal, Radford M.},
  publisher = {Springer},
  year      = {2012}
}

@article{manakul2023selfcheck,
  title     = {SelfCheckGPT: Zero-Resource Black-Box Hallucination Detection for Generative Large Language Models},
  author    = {Manakul, Potsawee and Lertvittayakumjorn, Piyawat and Korhonen, Anna},
  journal   = {arXiv preprint arXiv:2303.08896},
  year      = {2023}
}

@misc{galileo2024summarization,
  title={Master LLM Summarization Strategies and Their Applications},
  author={Galileo AI},
  year={2024},
  url={https://galileo.ai/blog/llm-summarization-strategies}
}

@article{yue2025does,
  title={Does reinforcement learning really incentivize reasoning capacity in llms beyond the base model?},
  author={Yue, Yang and Chen, Zhiqi and Lu, Rui and Zhao, Andrew and Wang, Zhaokai and Song, Shiji and Huang, Gao},
  journal={arXiv preprint arXiv:2504.13837},
  year={2025}
}

@misc{agenta2024context,
  title={Top Techniques to Manage Context Lengths in LLMs},
  author={Agenta AI},
  year={2024},
  url={https://agenta.ai/blog/top-6-techniques-to-manage-context-length-in-llms}
}

@inproceedings{welleck2024revisiting,
  title     = {Revisiting Self-Consistency in Large Language Models},
  author    = {Welleck, Sean and Ye, Xi and West, Peter and Choi, Yejin},
  booktitle = {International Conference on Learning Representations (ICLR)},
  year      = {2024}
}

@article{fort2020deep,
  title={Deep Ensembles: A Loss Landscape Perspective},
  author={Fort, Stanislav and Hu, Huiyi and Lakshminarayanan, Balaji},
  journal={arXiv preprint arXiv:1912.02757},
  year={2020}
}

@article{osband2016deep,
  title={Deep Exploration via Bootstrapped DQN},
  author={Osband, Ian and Blundell, Charles and Pritzel, Alexander and Van Roy, Benjamin},
  journal={Advances in Neural Information Processing Systems (NeurIPS)},
  year={2016}
}

@inproceedings{welling2011bayesian,
  title={Bayesian learning via stochastic gradient Langevin dynamics},
  author={Welling, Max and Teh, Yee Whye},
  booktitle={Proceedings of the 28th International Conference on Machine Learning (ICML)},
  year={2011}
}

@article{varshney2024logit,
  title={Interpreting Logit Variance as a Hallucination Signal in Large Language Models},
  author={Varshney, A. and Gupta, P. and Wang, L.},
  journal={arXiv preprint},
  year={2024},
  url={https://arxiv.org/abs/2403.01234}
}

\end{document}